\documentclass[runningheads]{llncs}

\usepackage{eccv}

\usepackage[bottom]{footmisc}

\usepackage{titletoc}
\usepackage{ifthen}
\newboolean{main}
\setboolean{main}{true}
\newboolean{suppmat}
\setboolean{suppmat}{true}

\newcommand{\titleTemplate}{%
A Probability-guided Sampler \\ for Neural Implicit Surface Rendering%
}
\newcommand{\titleRunning}{
A Probability-guided Sampler for Neural Implicit Surface Rendering}%
\newcommand{\authorsTemplate}{%
Gon\c{c}alo Dias Pais\textsuperscript{1,2}\thanks{Work was partly done while interning at MERL.}, Valter Piedade\textsuperscript{2}, Moitreya Chatterjee\textsuperscript{1}, \\ Marcus Greiff\textsuperscript{3}, and Pedro Miraldo\textsuperscript{1}
}
\newcommand{\authorRunning}{%
G. D. Pais, V. Piedade, M. Chatterjee, M. Greiff, and P. Miraldo
}
\newcommand{\instituteTemplate}{%
Mitsubishi Electric Research Laboratories (MERL), Cambridge\and
Instituto Superior T\'ecnico, Universidade de Lisboa \and
Toyota Research Institute, Los Altos %
}

\definecolor{BlueGrid}{HTML}{0057FF}
\definecolor{BluePart}{HTML}{70C0FF}
\usepackage{eccvabbrv}
\usepackage[accsupp]{axessibility} 
\usepackage{orcidlink}
\usepackage{preamble}
\usepackage{caption}

\begin{document}
\ifthenelse{\boolean{main}}
{
\startcontents[corpstexte]
\title{\titleTemplate} 
\titlerunning{\titleRunning}
\author{\authorsTemplate}
\authorrunning{\authorRunning}
\institute{\instituteTemplate}
\maketitle
\setcounter{footnote}{0}

\begin{abstract}
Several variants of Neural Radiance Fields (NeRFs) have significantly improved the accuracy of synthesized images and surface reconstruction of 3D scenes/objects. In all of these methods, a key characteristic is that none can train the neural network with every possible input data, specifically, every pixel and potential 3D point along the projection rays due to scalability issues.
While vanilla NeRFs uniformly sample both the image pixels and 3D points along the projection rays, some variants focus only on guiding the sampling of the 3D points along the projection rays. In this paper, we leverage the implicit surface representation of the foreground scene and model a probability density function in a 3D image projection space to achieve a more targeted sampling of the rays toward regions of interest, resulting in improved rendering.
Additionally, a new surface reconstruction loss is proposed for improved performance. This new loss fully explores the proposed 3D image projection space model and incorporates near-to-surface and empty space components. 
By integrating our novel sampling strategy and novel loss into current state-of-the-art neural implicit surface renderers, we achieve more accurate and detailed 3D reconstructions and improved image rendering, especially for the regions of interest in any given scene.  Project page: \url{https://merl.com/research/highlights/ps-neus}.

\end{abstract}
\section{Introduction}
Recovering the 3D structure of the scene and rendering it from new views is valuable for numerous tasks such as Augmented Reality/Virtual Reality asset creation, 3D reconstruction~\cite{wei2020deepsfm, wen2023bundlesdf}, environment mapping~\cite{Sandstrom_2023_ICCV, Kong_2023_CVPR, Zhu_2022_CVPR, Sucar_2021_ICCV}, etc.
In the last few years, Neural Radiance Fields (NeRF)~\cite{mildenhall2020nerf} have emerged as a promising solution for this task.
These learn a mapping from a 3D point and a viewing direction to its color and volume density. 
In theory, it may be desirable to train NeRFs on every pixel and every scene image from the training data. 
However, given the large amount of data, this is infeasible, \ie, %
one must sample the image pixels uniformly and the points on the projection ray.

Recent approaches like NeuS~\cite{wang2021neus} and its variants~\cite{wang2022neus2, wang2022hf, johnson2023unbiased, li2023neuralangelo} leverage neural implicit representations to achieve finer detail and higher-resolution 3D surface reconstruction, particularly of 3D objects. These methods typically employ Signed Distance Fields (SDF) or occupancy to implicitly represent the foreground surfaces. All NeuS-inspired methodologies adhere to a standardized sampling strategy involving uniform sampling of pixels and corresponding projection rays followed by hierarchical sampling of 3D points along the projection rays.

\begin{figure}[t]
\centering
{\scalebox{0.64}{\input{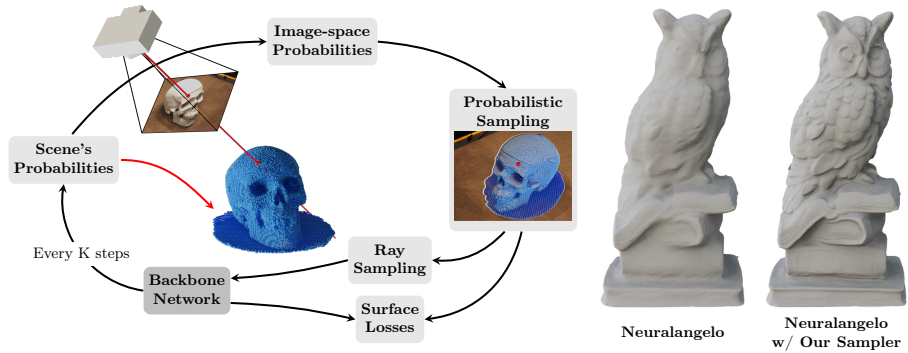}}}
\caption{{\bf Schematic of our pipeline and an example:}
We leverage neural implicit representations to guide the sampling of rays during the training of the neural surface rendering pipeline.
In particular, we sample a {\color{red} point} in the three-dimensional image space (image coordinates + depth) (\textbf{Probability-guided Sampling}) to obtain the ray (\textbf{Ray Sampling}) and some additional points around the surface, along the ray. These are then passed through the \textbf{Backbone Network}, which is regularized using the sampled depth (\textbf{Surface Reconstruction Losses}). The two figures on the right show a 3D reconstruction
of an example DTU scene (DTU122) using Neuralangelo as a backbone, with (rightmost) and without (second from right) our sampling strategy.
}
\label{fig:teaser}
\end{figure}

This paper argues that the full potential of neural implicit representations has not yet been explored. While many existing methods focus solely on guiding the sampling of 3D points along projection rays, disregarding the crucial aspect of pixel sampling~\cite{mildenhall2020nerf, barron2021mip, Barron_2022_CVPR, barron2023zip}, trivial solutions to pixel sampling, such as uniform sampling, generate worse rendering quality in areas of interest in the scene due to insufficient representation during training. This leads us to the following research question: \textit{Can the implicit surface representation guide training rays and points for accurate 3D reconstruction and rendering?}

There are two trivial solutions to the proposed research question: (i) The more straightforward one would be to cast a ray for every possible pixel in every camera and check whether any volume density accumulated along the ray. This solution is computationally expensive since one must run the model for all cameras and all pixels. Indeed, Sun \etal \cite{sun2024efficient} partially follows this idea but reduces the sampling space by evaluating it in patches to make the training feasible. The pixels are still uniformly sampled inside each patch.
(ii) The second approach voxelizes the implicit surface representation and projects every possible 3D point to every camera, which is computationally expensive. Additionally, occlusions -- which need to be factored in for view-dependent rendering -- cannot be trivially handled using such a technique. Similar ideas were followed in works such as RegSDF~\cite{Zhang_2022_CVPR}, where the authors use the Structure from Motion (SfM) points instead. However, this is a simplification of the trivial solution  
and only handles much simpler sampling scenarios, which do not need to deal with voxelization, view dependency, and training issues raised from dense 3D points.

Our paper introduces a novel solution for image pixel sampling. It leverages a 3D probability density function estimated from the scene's SDF within the 3D image space facilitated by per-camera grids. This enables efficient interpolation of camera pixels and depth approximation. 
Our method offers a streamlined update process, requiring only a single model run for the 3D probability density function update, thus ensuring speed and adaptability. Unlike conventional methods relying on additional depth data, our approach constructs the sampling space through 3D coordinate transformations and view constraints without any additional depth supervision. Notably, our technique enhances 3D scene rendering across various backbones.
The sampling pipeline and a mesh reconstruction are represented in~\cref{fig:teaser}.
Our main contributions are:
\begin{enumerate}
    \item A probabilistic 3D orthographic image projection sampling for neural implicit surface rendering that is view-dependent and feasible for training;
    \item A new loss function that combines near-the-surface and empty space components for better modeling foreground and background regions of the scene;
    \item The sampling is agnostic to the implicit model, \ie, the derived extra pipeline steps can be used with different models without changing the backbone;%
    \item We show that coupling our probabilistic sampling with current state-of-the-art neural implicit representation methods (namely~\cite{wang2021neus, li2023neuralangelo}) improves 3D reconstruction and rendering in regions of interest of the scene.
\end{enumerate}

\section{Related Work}\label{sec:related}

\paperpar{Neural rendering}
Reconstructing 3D structures from multi-view images is a core problem of computer vision, with approaches based on SfM~\cite{Schonberger_2016_CVPR, schonberger2016pixelwise, wei2020deepsfm, geppert2020privacy} or Simultaneous Localization and Mapping (SLAM)~\cite{davison2003real, campos2021orb, Sandstrom_2023_ICCV, Kong_2023_CVPR, Zhu_2022_CVPR, Sucar_2021_ICCV}. %
NeRF~\cite{mildenhall2020nerf} introduces a more recent view synthesis strategy that enables dense reconstructions, using volume rendering. During training, projection rays and 3D points are uniformly sampled from the image and any given projection ray, respectively. Image synthesis is achieved by rendering the sampled 3D points by volume, which gives their color and volume density values. Lately, several NeRF variants have been proposed, focusing on improving view synthesis quality~\cite{barron2021mip, kaizhang2020, Barron_2022_CVPR}, improving computational performance~\cite{muller2022instant, barron2021mip, sun2022direct}, scaling up to large-scale environments~\cite{Tancik_2022_CVPR, turki2022mega, Deng_2023_ICCV, Zhang_2022_CVPR_1}, dynamic scenes~\cite{Tretschk_2021_ICCV, Park_2021_ICCV, Pumarola_2021_CVPR, Li_2022_CVPR, Li_2021_CVPR, attal2023hyperreel, liu2024gear}, \etc. 

\paperpar{Neural implicit volume rendering}
A drawback of the volume rendering in NeRF is that it imposes insufficient constraints for representing 3D surfaces. This prevents it from learning intricate 3D object details, making high-quality reconstructions infeasible. To solve this problem, occupancy approaches~\cite{oechsle2021unisurf, peng2020convolutional, mescheder2019occupancy, Niemeyer_2020_CVPR, saito2020pifuhd, liu2020neural} and SDF approaches~\cite{zhang2021learning, wang2021neus, Zhang_2022_CVPR, wang2022neus2, li2023neuralangelo, sun2021neuralrecon, azinovic2022neural, yariv2023bakedsdf, darmon2022improving, yariv2021volume, zhang2023towards, gaur2024orientedgrid} were proposed. %
In particular, NeuS~\cite{wang2021neus} and VolSDF~\cite{yariv2021volume} use SDF as an implicit representation for a surface and are trained from multiple views.
Both outperform NeRF-based methods, even handling scenes with occlusions. Further improvement was achieved by reducing the implicit bias, in the depth estimates~\cite{zhang2023towards}. Another issue with NeuS is its slow training speed. NeuS2~\cite{wang2022neus2} proposes an efficient parallelization and a new training strategy to address this concern. 
Neuralangelo~\cite{li2023neuralangelo} introduces multi-resolution hash grids for surface rendering. This approach achieves high-quality reconstruction in highly detailed scenes, with a small cost in training efficiency. While these approaches use the learned implicit representation for sampling on the projection rays, our work utilizes a probability density function for sampling. 
To further improve surface reconstruction, other approaches use priors such as object masks~\cite{Niemeyer_2020_CVPR, yariv2020multiview}, depth~\cite{zhang2021learning, yu2022monosdf}, normals~\cite{wang2022neuris, yu2022monosdf}, or point clouds~\cite{Zhang_2022_CVPR, fu2022geo}. These additional inputs guide the surface learning process, improving the reconstruction results and optimization time.
We propose a method where sampling is guided by surface estimates, enabling rays to converge toward textured regions without the need for additional inputs.

\paperpar{Pixel Sampler for NeRF} The straightforward approach to sampling the rays while training NeRFs is to uniformly sample both the pixels in the image as well as the 3D points that lie on the corresponding projection rays, passing through the chosen pixels~\cite{mildenhall2020nerf}.
Most sampling strategies in NeRF propose to improve the 3D sampling on the ray,
by exploring anti-aliasing~\cite{barron2021mip, kurz2022adanerf, barron2023zip, hu2023tri} or 3D geometry~\cite{liu2020neural, wang2021neus, yariv2021volume}. In contrast, this paper focuses on a more effective pixel sampling strategy for training similar to~\cite{sun2022neural, wu2023actray, sun2024efficient}.
Sun \etal~\cite{sun2024efficient} proposes a patch sampling based on the depth and color contrast estimates for their pixel sampling, where a pre-trained model trained on the DTU~\cite{jensen2014large} is used to obtain the initial proposals. Neural 3D reconstruction in the Wild~\cite{sun2022neural} attempts to sample only around the surface through voxel-guided and surface-guided sampling. ActRay\cite{wu2023actray} uses a reinforcement learning agent to reduce the number of rays by focusing on the rays with the highest loss values.
In contrast, we design a 3D view-dependent camera probability space, derived from the implicit representation of the surface to sample the pixel and directly gain depth information for the sampled ray. A backbone model, upon which our sampling strategy operates, is trained from scratch  
without needing additional information.

\section{Notations and Background}\label{sec:background}

\subsection{Notations}\label{subsection:basics}
Let a 3D point in world coordinates be given by $\xvec = [x, y, z] \in \mathcal{X}\subset\mathbb{R}^3$. For a set of cameras $\Ccal = \{1, \dots, C\}$, the same point in the $c$\textsuperscript{th} camera, %
is denoted by $\widehat \xvec_c = [\widehat x_c, \widehat y_c, \widehat z_c] \in \widehat{\mathcal{X}}_c$. $h_c(\cdot)$ transforms the point from the world to the camera $c$ (see \cite{Hartley2004}). For $c$, we define a {\it 3-dimensional image space} such that
\begin{equation}
\uvec_c \in \mathcal{U}_c = g(\widehat \xvec_c) = [\widehat x_c / \widehat z_c, \widehat y_c / \widehat z_c, \widehat z_c] = [u_c, v_c, \lambda_c],
\label{eq:3d_image_space_definition}
\end{equation}
where $u_c$, $v_c$, and $\mathcal{U}_c$ (respectively) are bounded to image size and intrinsic parameters of each camera, and depth $\lambda_c > 0$, which is bijective to the camera reference frame\footnote{Proof and more details given in the supplementary material.}.
Using the transformation from the world to the camera coordinate system $h_c(\cdot)$ and image projection $g(\cdot)$ (for more detail see \cite{Hartley2004}), we define the composition $f_c(\cdot)$, such that
\begin{equation}\label{eq:camera_projection}
\uvec_c = f_c(\xvec) = g(h_c(\xvec)).
\end{equation}
To simplify the notations, we sometimes omit the subscript $c$, for example, $\uvec = \uvec_c$, and $\widehat \xvec = \widehat \xvec_c$. Finally, $|\cdot|$ denotes the determinant of a matrix.

\subsection{Neural Implicit Surface Rendering}\label{subsec:neural_implicit_surface_rendering}

Consider a set of images of a specific 3D scene, captured from calibrated cameras with known poses. A NeRF~\cite{mildenhall2020nerf} creates an implicit 3D representation of the scene from known camera positions to the images. %
This implicit representation allows for a dense reconstruction of the scene, by simultaneously estimating the volume density and color for every 3D point.
A more evolved alternative is proposed in Wang \etal~\cite{wang2021neus}. The authors introduce a novel approach to estimate densities from an SDF representation by approximating it using a logistic function: %
\begin{equation}
\phi_s(o) = (se^{-so})/(1 + e^{-so}),
\label{eq:logistic_fuction}
\end{equation}
where $s$ is the logistic scale and $o$ the SDF output.
This conversion enables the application of camera-free volume rendering techniques for scene reconstruction. Using the SDF, the scene's outer surface $\mathcal{S}$ is represented as the zero-level set, defined as $\mathcal{S} = \{ \xvec \in \mathbb{R}^3: S(\xvec) = 0 \}$, where $S(\cdot)$ is the output of the SDF network. The rendering is then computed using the SDF at a particular 3D point. The volume density at each point along the ray is
\begin{equation}\label{eq:opacity}
\alpha_i = \max \left( \frac{\Phi_s(S(\xvec_i)) - \Phi_s(S(\xvec_{i+1}))}{\Phi_s(S(\xvec_i))}, 0 \right),
\end{equation}
where $\Phi_s(\cdot)$ is the sigmoid function\footnote{$\phi_s(\cdot)$ is the derivative of $\Phi_s(\cdot)$, hence $\Phi_s(\cdot)$ is the cumulative density function of the logistic distribution.}. The accumulated volume density is
\begin{equation}\label{eq:rendering_weights}
    w_i = \alpha_i T_i = \alpha_i \prod_{j=0}^{i} (1 - \alpha_j),
\end{equation}
where $T_i$ is the transmittance at the point $i$ along the ray.
See \cite{wang2021neus} for details.

\begin{figure}[t]
    \centering
    \scalebox{0.68}{ \input{figs/pipeline} }
    \caption{{\bf Proposed guided-sampling:} %
    The scene is represented as a 3D grid $\Gcal_\Xcal$, and characterized by a PDF $p(\xvec)$ computed from the SDF network and modeled by a logistic distribution of the SDF values $\phi_s(S(\xvec))$. 
    We propose to use a 3D image space that includes depth, represented as $\Gcal_\Ucal$, where one can define $p(\uvec)$ based on $p(\xvec)$
    as described in \textbf{Interpolation} -- \cref{subsec:interpolation}. Then, we consider the camera viewpoint of the scene (such as occlusions), by weighting $p(\uvec)$ as described in \textbf{View Dependency} -- \cref{subsec:view_dependency}. %
    In the shown grids, {\color{blue} color hue} maps to the probability value, normalized for each grid. A higher hue is more probable.
    At every training step, points are sampled from $\widetilde{p}(\uvec)$ to create {\color{red} ray samples} $\tilde \uvec$ (\textbf{Probabilistic Sampling} -- \cref{subsec:prob_sampling}).
    We sample {\color{ForestGreen} rays uniformly} to allow overall image quality and scene exploration.
    }
    \label{fig:neus_pipeline}
\end{figure}

\section{Proposed Approach}\label{sec:method}
This work introduces a new probability-guided sampler for enhanced scene rendering and 3D reconstruction, seamlessly merging with neural surface pipelines.

\subsection{Method Overview}\label{subsec:method_overview}
Consider a typical neural surface rendering pipeline, such as the one proposed by NeuS~\cite{wang2021neus}. An intermediate step of such methods consists of obtaining an SDF that models the 3D structure of the foreground of a scene  (see {\bf Neural Surface Rendering Pipeline} block in \cref{fig:neus_pipeline}). In this work, we intend to utilize the SDF for more effective sampling while training the neural volume field. In particular, we intend to focus on the important regions of the scene, \ie foreground, when training.
Additionally, we leverage the output information from our sampling module to aid the sampling process along the rays and the training with additional surface reconstruction losses. Our method does not require any additional information (\eg SfM points) or models.
The proposed training pipeline is depicted in \cref{fig:neus_pipeline}, {\bf Probability-guided Sampler}.

We start by leveraging SDF representation in \cref{eq:logistic_fuction} to define a Probability Density Function (PDF) over the points in the 3D scene to capture the likelihood of it being sampled during training, denoted as $p(\xvec)$:
\begin{equation}\label{eq:scene_prob}
p(\xvec) = \phi_s\left(S(\xvec)\right).
\end{equation}
Then, we explore a suitable 3D image space from $p(\xvec)$ for effective sampling in the camera's viewpoint.
To compute the probability in a 3D image space, the transformation has to be bijective and consequently invertible to account for the change of variables. 
From a geometric point of view, the proposed space $\Ucal$ is obtained from $\Xcal$ by transforming the projection rays, which, by definition, are parallel to each other and perpendicular to the image space (\ie orthographic projection space). The new PDF is $p(\uvec)$ and is described in~\cref{subsec:interpolation}.

Next, we deal with the concerns arising from view dependency, such as occlusions. Rather than sampling directly in the image from the scene's projection and probabilities, where awareness of the viewpoint is limited, we weigh the camera's PDF $p(\uvec)$ using a volume rendering strategy. This allows for seamless integration of view dependency constraints and provides the foundation for the sampling process. In \cref{subsec:view_dependency}, this PDF is defined as $\widetilde{p}(\uvec)$.

The final step of our formulation consists of sampling 3D points on ${\Ucal}$ using $\widetilde{p}(\uvec)$. The proposed method follows a conditional sampling strategy detailed in~\cref{subsec:prob_sampling}. \cref{subsec:surflosses} describes how the sampled $\uvec$ is used in the neural surface pipeline and details the proposed surface regularization losses designed to guide the training process by considering the sampled depth.

\subsection{Interpolation}\label{subsec:interpolation}

We aim to transform the density estimate $p(\xvec)$, for a point $\xvec \in \Xcal$ to the 3-dimensional image space, defined in \cref{eq:3d_image_space_definition}, which is denoted by $p(\vvec)$, where $\vvec \in \Ucal$. This transform is given by
\begin{equation}\label{eq:change_coord}
p\left(\xvec\right) = p\left(f^{-1}\left(\vvec\right)\right)\left|\frac{\partial f^{-1}\left(\vvec\right)}{\partial\vvec}\right| \implies
p\left(f^{-1}\left(\vvec\right)\right) = \eta^{-2} p\left(\xvec\right),
\end{equation}
where $\vvec = f(\xvec)$ and $\eta$ the depth value of $\vvec$.

\begin{SCfigure}[2][t]
    \scalebox{1}{\input{figs/interpolation_fig}} 
    \caption{\textbf{Bird's-eye view illustration of the Riemann integral approximation of $p(\uvec)$:} This figure shows the 3D image space in {\color{red} red}, the scene grid transformed to the image space in {\color{BlueGrid} blue} and how probability of ${\color{orange}\uvec} \in  \gridU$ ({\color{orange} orange}) is computed. A {\color{BlueGrid}projected scene point \faCircle } $\in \Gcal_{f(\Xcal)}$ that lies inside the {\color{orange} cell \faSquareO} will contribute to the computation of $p(\uvec)$, as shown in the equation.%
    }
    \label{fig:interpolation_scheme}
\end{SCfigure}

To simplify and have a more compact representation, we discretize the 3D scene space $\xvec$, such that $\xvec \in \gridX \subset \Xcal$, and the 3D image space of $\uvec$ such that $\uvec \in \gridU \subset \Ucal$, with $\gridX$ and $\gridU$ denoting a discretized grid on $\Xcal$ and $\Ucal$. 
Note that $\vvec$ is not represented in the grid of the 3-dimensional image space $\gridU$. Instead, $\vvec$ is discretized according to the scene grid $\gridX$ after applying the camera transformation $f(\cdot)$ in \cref{eq:camera_projection}, which we define as $\Gcal_{f(\Xcal)}$. However, the probability estimates $p(\uvec)$ in $\gridU$ cannot be easily interpolated from~\cref{eq:change_coord} due to the respective space deformation resulting from the discretization and transformation. 
Therefore, we approximate $p(\uvec)$ as the Riemann integral of all transformed cells of $\gridX$ in $\uvec$. We start by making sure $\gridX$ is discretized finely\footnote{To make sure that $\gridX$ is small enough, the scene $\gridX$ is partitioned by a factor $F$, where each cell is divided at each axis into $F$ equal parts, resulting in a total of $F^3$ equal cells from the $3$ axes. The probability of the newly partitioned cell is the original cell probability divided by $F^3$ (see supplementary material for more details).}. 
For each $\uvec \in \gridU$, the probability estimate is the sum of the probability densities of all points $\Gcal_{f(\Xcal)}$ that lie inside the cell, as defined in~\cref{eq:change_coord}. A depiction of the interpolation of $p(\uvec)$ is shown in~\cref{fig:interpolation_scheme}, where the selected transformed cells have an orange fill. Further details in supplementary material.

\subsection{View Dependency}\label{subsec:view_dependency}
Since $p(\uvec)$ does not account for occlusions created by the camera's perspective projection, sampling a projection ray %
based on the object's geometry alone can result in too many occluded samples and, consequently, loss of training efficiency. 
To address this issue, we assume that the volume density $\sigma$ per cell is $p(\uvec)$ as a naive solution discussed in Wang \etal~\cite{wang2021neus}\footnote{Note that this density representation along the ray causes a bias in the depth estimate~\cite{wang2021neus, zhang2023towards}. Nonetheless, this formulation remains occlusion-aware.}. The transmittance $T$ can then be evaluated in the 3-dimensional image space by accumulating the radiance weighted by the volume densities for cells along the ray, corresponding to the image coordinates $[u, v]$. Considering the grid $\gridU$, the transmittance $T_i$ at the depth $\lambda_i$ corresponding to the $i$-th cell along $[u, v]$ can be defined as $T_i = e^{-\sum_{k = 1}^{i} p([u, v, \lambda_k]^T)}$, where the $k^{th}$-cell is sorted by depth.
Then, the view-dependent probability $\widetilde p (\uvec_i)$ for $\uvec_i = [u, v, \lambda_i]^T$ is defined as the transmittance weighted by the volume density accumulated along a ray, as shown in~\cref{fig:neus_pipeline},
\begin{equation}
    \widetilde p(\uvec_i) = \sigma_i T_i = p(\uvec_i) e^{-\sum_{k = 0}^{i} p([u, v, \lambda_k]^T)}.
\end{equation}

\subsection{Probability-guided Sampling}\label{subsec:prob_sampling}
While in previous neural rendering pipelines, pixels are sampled in the image uniformly, in this work, we combine the two sampling strategies:
(i) sampling using the view-dependent space PDF $\widetilde p(\uvec)$, and (ii) sampling uniformly on the image. The former is better suited for the foreground and the latter for the background, allowing us to regulate the proportion of samples around the image.

Starting with the view-dependent space sampling, we use conditional probabilities, which extend ray importance sampling in~\cite{pharr2023physically} to the 3-dimensional space $\Ucal$. %
The first marginal density function is then defined as
\begin{equation}\label{eq:first_marginal}
    \widetilde p(u) = \frac{1}{R_v R_\lambda} \sum_{(v, \lambda)} \widetilde p([u, v, \lambda]^T),
\end{equation}
for all $(v, \lambda)$ cells of $u$, where $R_v$ and $R_\lambda$ are resolutions of the grid $\gridU$ along the axes of $v$ and $\lambda$. Then, the first conditional distribution is computed as
\begin{equation}
    \widetilde p(v, \lambda | u) = \frac{\widetilde p(\uvec)}{\widetilde p(u)}.%
\end{equation}
The second marginal applied to $v$ can then be expressed as
\begin{equation}\label{eq:second_marginal}
    \widetilde p(v | u) = \frac{1}{R_\lambda} \sum_{\lambda} \widetilde p(v, \lambda | u),
\end{equation}
for all $\lambda$ cells. 
Finally, the second conditional distribution is then defined as
\begin{equation}\label{eq:second_conditional}
    \widetilde p(\lambda | u, v) = \frac{\widetilde{p}  (v, \lambda | u)}{\widetilde p(v | u)}.
\end{equation}

With the marginals and conditionals defined, we sample $\widetilde \uvec = [\widetilde u, \widetilde v, \widetilde \lambda] \in \Ucal$ in the 3-dimensional image space, to obtain the 3D projection ray and image pixels. We start by sampling $\widetilde u$ from the first marginal, \cref{eq:first_marginal}, using inverse transform sampling~\cite{mosegaard1995monte}.
Then, we approximate the second marginal $p(v | \widetilde u)$ from the samples $\widetilde u$ using bilinear interpolation. Following the same inverse sampling strategy, $\widetilde v$ is sampled according to $p(v | \widetilde u)$. Finally, using trilinear interpolation, we approximate the second conditional $p(\lambda | \widetilde u, \widetilde u)$, with $\widetilde u$ and $\widetilde v$, and sample $\widetilde \lambda$. 

When sampling uniformly along the rays during training and evaluation, we interpolate the second conditional, \cref{eq:second_conditional}, with given values for $\widetilde u$ and $\widetilde v$, and sample $\widetilde \lambda$ directly. A representation of the sampled output is depicted in~\cref{fig:neus_pipeline}.

\subsection{Surface Reconstruction Losses} %
\label{subsec:surflosses}

The input to the rendering network (irrespective of what backbone is used) is the 3D points sampled along the rays from the sampled pixel obtained from $\widetilde \uvec$.
In addition to these obtained by following the sampling strategy of the backbone network, we provide additional 3D points along the ray near the sampled depth $\widetilde \lambda$ for improved rendering of such regions, keeping the same sampling budget. This is accomplished by drawing samples from a Gaussian distribution $\sim \mathcal{N}(\tilde{\lambda}, \frac{\pi^2}{3 s^2})$, where the variance is determined by the normal approximation of the logistic distribution~\cite{stefanski1991normal}, with the mean being the sampled $\tilde{\lambda}$. 

In addition to each backbone, we introduce losses for points near the surface (near zero-level set), points within the empty ray space, and points belonging to background rays. 
Consider $M$ projection rays and $N_{\text{fg}}$ foreground points sampled along those rays. The proposed near-surface loss accounts for sampled points within 99.7\% of the possible near-surface samples during ray sampling, \ie, points for the $m$-th ray, where $m \in \{1, \dots, M\}$, satisfying $\mathcal{N}_{\text{Near}} = \{f(\xvec) \in \Ucal: f(\xvec) \in [(u, v, \tilde{\lambda} - 3 \frac{\pi}{\sqrt{3} s}), (u, v, \tilde{\lambda} + 3 \frac{\pi}{\sqrt{3} s})] \}$, and is given by
\begin{equation}
L^{\text{Near}}_m = \sum_{i \in \mathcal{N}_{\text{Near}}} |S(\xvec_i)| w_i,
\end{equation}
where $S(\cdot)$ is the SDF value, and $w_i$ represents the volume density accumulated along a ray of the point $i$, given by~\cref{eq:rendering_weights}.

For points in the empty ray space, \ie, the complement set of $\mathcal{N}_{\text{Near}}$, denoted as $\mathcal{N}_{\text{Empty}}$, we introduce a loss to encourage small SDF values and exploration:
\begin{equation}
L^{\text{Empty}}_m = \sum_{j \in \mathcal{N}_{\text{Empty}}} \left[(S(\xvec_j) - \epsilon) w_j \right]^2,
\end{equation}
where $\epsilon$ is a small value.
We consider view dependency in both losses by incorporating the accumulated volume densities, $w_a$, where $a \in \{i, j\}$.

Finally, for rays that do not intersect foreground surfaces, \ie, if the sampled depth $\widetilde \lambda$ is outside of the scene's boundary, the following background loss ensures that the importance of accurately estimating the scene geometry decreases as one moves farther from the surface:
\begin{equation}
L^{\text{Bg}}_m = \sum_{k = 1}^{N_{\text{fg}}} e^{-\beta|S(\xvec_i)|} w_k.
\end{equation}

All losses are averaged by over $M$ rays.
The total surface loss is computed as $L^{\text{Surf}} = \lambda_1 L^{\text{Near}} + \lambda_2 (L^{\text{Empty}} + L^{\text{Bg}})$. This surface loss is appropriately weighted and added to the existing losses for each backbone.
\section{Experiments}\label{sec:experiments}

\subsection{Experimental Setup}\label{subsec:experimental_setup}

We use NeuS~\cite{wang2021neus} and Neuralangelo~\cite{li2023neuralangelo} as backbones to evaluate the effectiveness of our proposed approach. Specifically, we use Neuralangelo's official implementation\footnote{\href{https://github.com/NVlabs/neuralangelo}{https://github.com/NVlabs/neuralangelo}} with the default settings and implemented NeuS within Neuralangelo's framework. Also, we use the default settings for both models when evaluating the proposed probabilistic sampling. With respect to resolutions of the scene space and camera spaces, we use $128$ for the 3 dimensions of $\gridX$, and $R_{u} = R_{v} = 64$ and $R_{\lambda} = 128$, for the 3D image space $\gridU$. We set $\lambda_1 = \lambda_2 = 0.5$. We use the scene scale and centers provided with the dataset to define each scene boundary. For each camera, we utilize the image corners to obtain $\Ucal$ boundaries.
The depth boundary is computed by shooting a ray from the central image pixel and computing where the ray hits the scene's boundaries.
During training, we update the 3D image space every 2500 and 5000 iterations for Neuralangelo and NeuS, respectively. For a fair comparison, we use the same number of rays and ray points as the baselines.
We initialize the camera grids as a sphere~\cite{yariv2020multiview}, facilitating the initial scene exploration. No depth ground-truth is used to supervise or evaluate the model. 
At the start of training, we set 20\% of the rays to be sampled uniformly in the image. As training progresses, the percentage increases to 40\%, 60\%, and 80\%. 
In the rendering pipeline, we sample $32$ points around the sampled depth, as described in~\cref{subsec:surflosses}. We train all backbones with their default losses, with $L^{\text{Surf}}$ being assigned a weight of $500$. We train and test all models in an A40 GPU using $10$ CPU cores.

\paperpar{Datasets}
We use DTU~\cite{jensen2014large}, as the primary dataset, which has $15$ object-focused sequences in a controlled environment, with object masks available to evaluate intricate details. Each DTU scan has $49$ or $64$ views each. We also evaluate more diverse sequences in BMVS~\cite{yao2020blendedmvs} dataset, where we choose $5$ object-centric sequences and $4$ large-scale sequences from Tanks and Temples (TNT)~\cite{Knapitsch2017}. The last two datasets have around $200$ images in each sequence.

\begin{figure}[t]
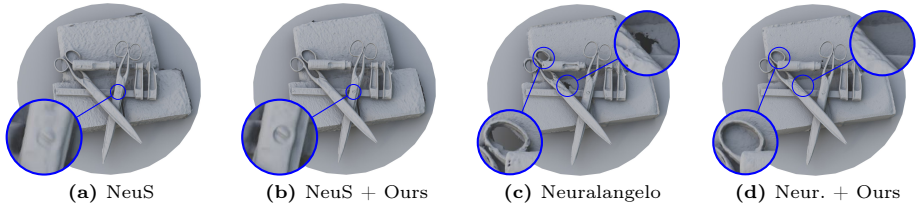

    \centering
    \begin{subfigure}{0.23\linewidth}
        \ZoomPictureB{figs/eccv/neus_dtu37}{0.55}{0.77}{}{5}
        \label{subfig:dtu37_neus}
        \caption{NeuS}
    \end{subfigure} \hfill
    \begin{subfigure}{0.23\linewidth} 
        \ZoomPictureB{figs/eccv/neus_w_sampler_dtu37}{0.55}{0.77}{}{5}
        \label{subfig:dtu37_neus_w_sampler}
        \caption{NeuS + Ours}
    \end{subfigure} \hfill
    \begin{subfigure}{0.23\linewidth}
        \ZoomPictureC{figs/eccv/neur_dtu37}{0.31}{1.10}{}{3}{0.44}{0.84}
        \label{subfig:dtu37_neur}
        \caption{Neuralangelo}
    \end{subfigure} \hfill
    \begin{subfigure}{0.23\linewidth}
        \ZoomPictureC{figs/eccv/neur_w_sampler_dtu37}{0.31}{1.10}{}{3}{0.44}{0.84}
        \label{subfig:dtu37_neur_ours}
        \caption{Neur. + Ours}
    \end{subfigure}
    \caption{{\bf DTU scan 37}: When our sampling and surface reconstruction losses are included in NeuS and Neuralangelo backbones, we get a sharper 3D reconstruction of the foreground objects. Our approach also removes the hole obtained by Neuralangelo.
    }
    \label{fig:dtu37_baselines}
\end{figure}
\begin{figure}[t]
    \centering
    \begin{subfigure}{0.48\columnwidth}
    \ZoomPictureD{figs/eccv/neur_dtu118_render_normal_crop}{0.55}{2.55}{}{4}
    \hspace{-20pt}\includegraphics[width=0.44\linewidth]{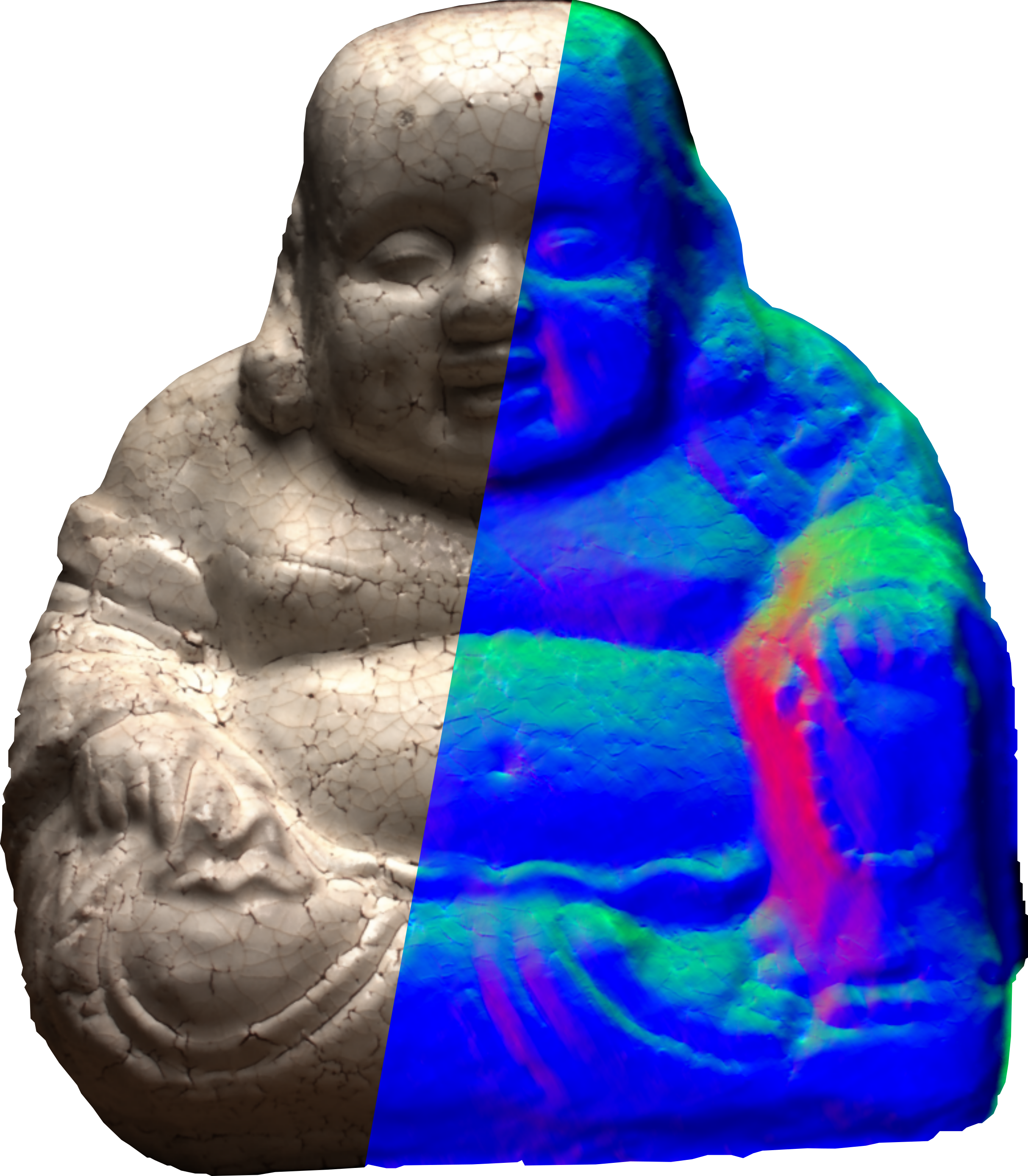}
    \caption{Neuralangelo}
    \end{subfigure}
    \hfill
    \begin{subfigure}{0.48\columnwidth}
    \ZoomPictureD{figs/eccv/neur_w_sampler_dtu118_render_normal_crop}{0.55}{2.55}{}{4}
    \hspace{-20pt}\includegraphics[width=0.44\linewidth]{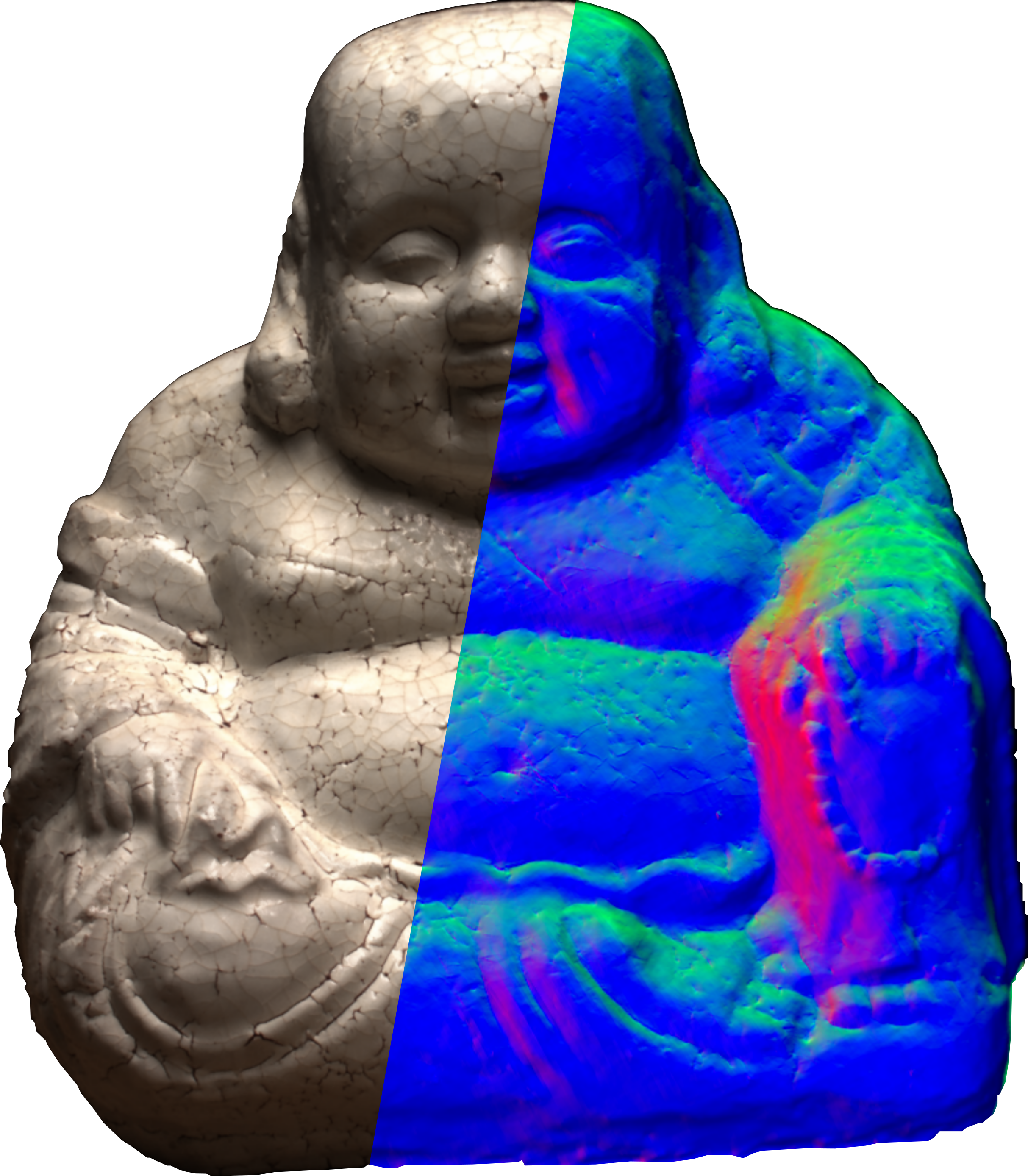}
    \caption{Neuralangelo~$+$~Ours}
    \end{subfigure}
    \caption{{\bf DTU scans 118 and 114}: We can extract more detailed meshes with our sampler and losses. The synthesized normal images are less noisy, with sharper edges in both scans, leading to better reconstruction. Image quality remains similar.}
    \label{fig:dtu118_baselines}
\end{figure}

\paperpar{Evaluation Metrics}
For image synthesis, we use the Peak Signal-to-Noise Ratio (PSNR). Following the evaluation protocols of prior work, we use NeuralWarp's evaluation methodology~\cite{darmon2022improving} for 3D reconstruction and report the Chamfer distance~\cite{jensen2014large} in DTU and F1-score~\cite{Knapitsch2017} in TNT. Note that we evaluate the models with object masks, when available, to measure the effectiveness of reconstructing the foreground. However, we do not use them during training.

\begin{figure}[t]
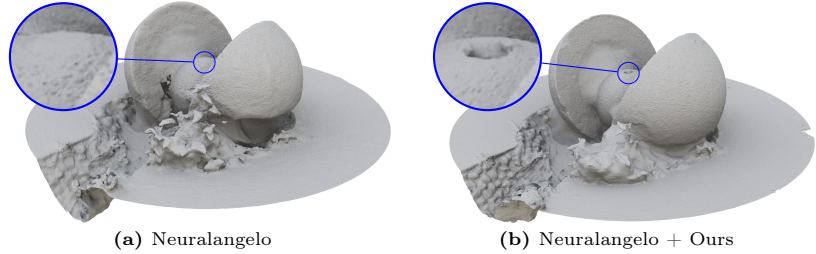

\centering
\begin{subfigure}{0.4\columnwidth}
\ZoomPictureA{figs/eccv/neur_bmvs_sphere}{0.49}{0.75}{}{5}
\caption{Neuralangelo}
\end{subfigure}
\qquad
\begin{subfigure}{0.4\columnwidth}
\ZoomPictureA{figs/eccv/neur_w_sampler_bmvs_sphere}{0.49}{0.70}{}{5}
\caption{Neuralangelo + Ours}
\end{subfigure}
\caption{\textbf{BMVS sequence "Sphere":} We observe that Neuralangelo + Ours get a significantly more complete 3D surface reconstruction. Also, our sampling better captures small and intricate regions, such as the small hole highlighted in the object foreground.}
\label{fig:bmvs_sphere}
\end{figure}

\paperpar{Baselines}
Our primary goal is to evaluate the impact of the probabilistic guided ray sampling by comparing it against approaches that do not use it. %
Towards this end, the main baselines are NeuS~\cite{wang2021neus}\footnote{Our implementation within Neuralangelo framework.} and Neuralangelo~\cite{li2023neuralangelo}, where we augment the proposed sampling strategy. For assessing the effectiveness of image synthesis and reconstruction quantitatively, we compare against existing literature: VolSDF~\cite{yariv2021volume}, RegSDF~\cite{Zhang_2022_CVPR}, NeuralWarp~\cite{darmon2022improving}, and HF-NeuS\cite{wang2022hf}.

\subsection{Results}
\label{subsec:results}
We discuss qualitative and quantitative results on the proposed sampler. We also comment on the ablation study and the computational overhead. %

\begin{figure}[t]
    \centering
    \begin{subfigure}{0.49\columnwidth}
        \includegraphics[width=0.49\linewidth]{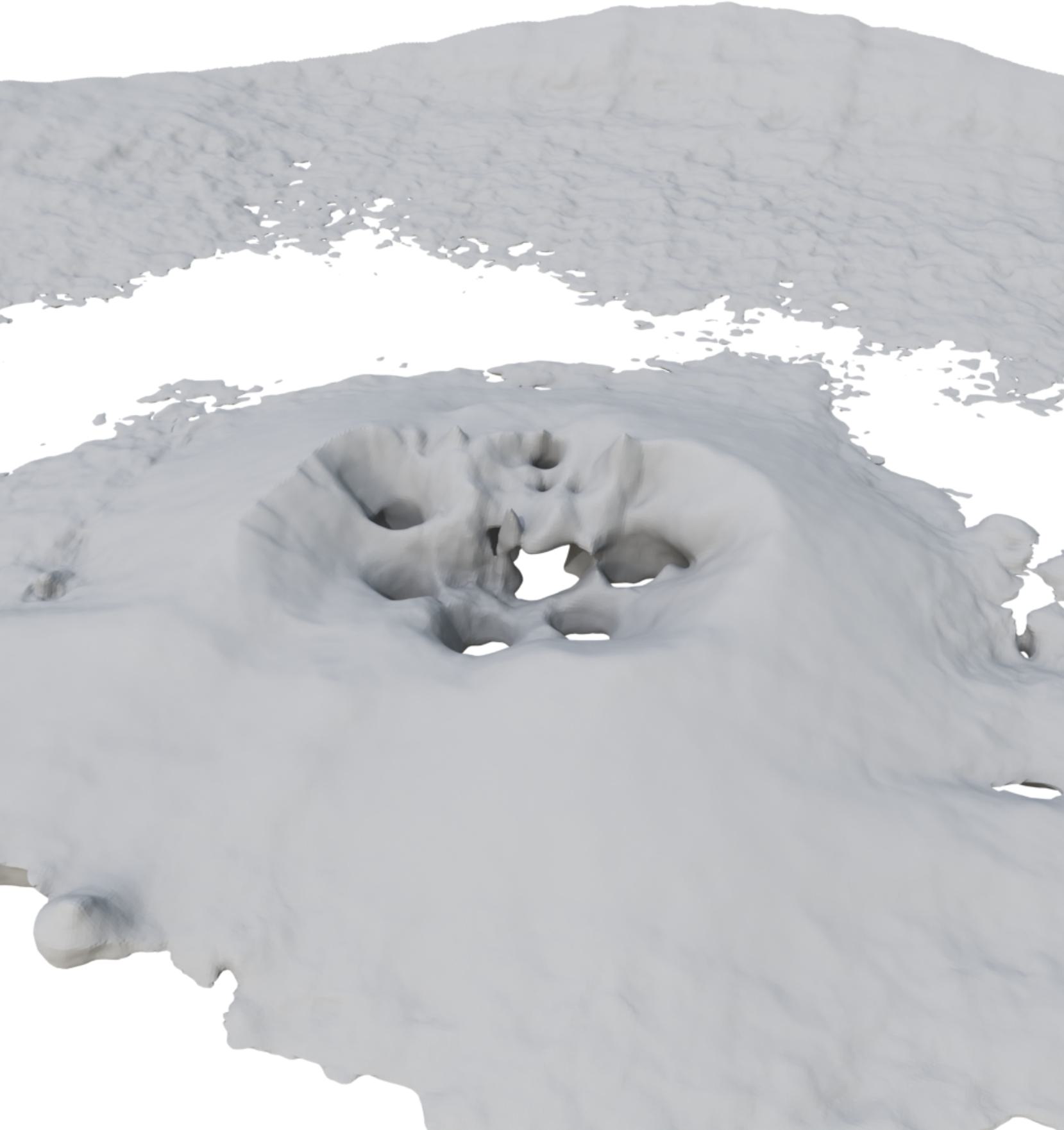}\ 
        \includegraphics[width=0.49\linewidth]{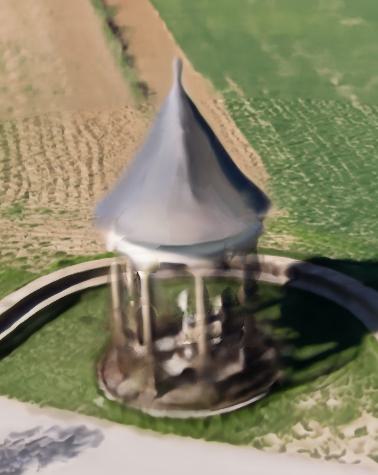}
        \caption{NeuS}
    \end{subfigure}
    \hfill
    \begin{subfigure}{0.49\columnwidth}
        \includegraphics[width=0.49\linewidth]{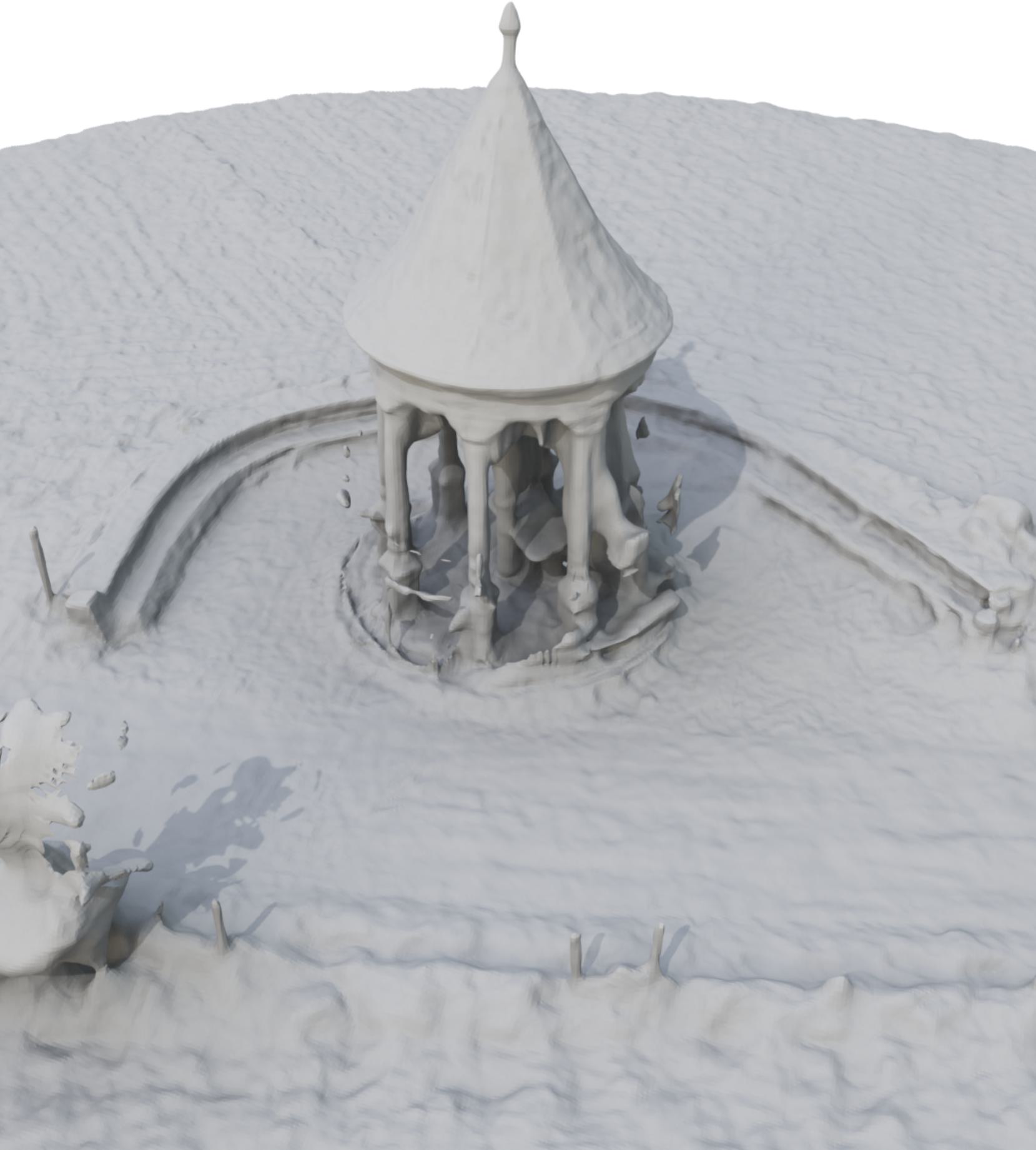}\ 
        \includegraphics[width=0.49\linewidth]{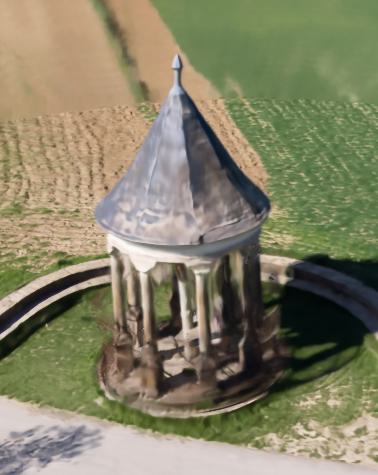}
        \caption{NeuS + Ours}
    \end{subfigure}
    \caption{\textbf{BMVS sequence "Bandstand"}: Our sampling and surface reconstruction losses have clear advantages. While NeuS fails to capture the foreground surfaces, the proposed sampling obtains a reasonable 3D structure and high-quality images.}
    \label{fig:bmvs_bandstand}
\end{figure}

\begin{SCfigure}[60][t]
    \centering
    \scalebox{1}{\includegraphics[width=0.4\linewidth]{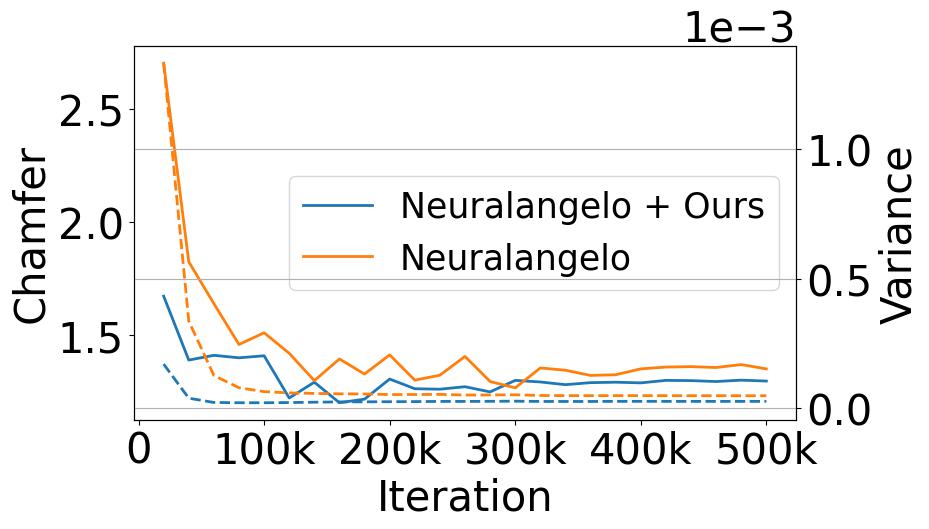}}
    \caption{\textbf{Reconstruction Error and Variance over training.} Our sampling converges faster to a better 3D representation (solid line), while significantly reducing the variance of the logistic output density (dashed).}
    \label{fig:s_var_}
\end{SCfigure}

\paperpar{Qualitative results}
We show a qualitative comparison of the DTU and BMVS datasets.
When observing the 3D reconstruction, the proposed approach preserves finer details, has fewer artifacts, increased completeness, and fewer holes as illustrated in~\cref{fig:teaser,fig:dtu37_baselines,fig:dtu118_baselines,fig:bmvs_sphere}. Even in complex scenes, the sampler reduces failure cases, such as in~\cref{fig:bmvs_bandstand}. Guiding the pixel sampling towards surface areas enhances 3D consistency and subsequently improves sharpness while displaying faster convergence and lower model variance as shown in~\cref{fig:s_var_}. See supplementary material for more qualitative results, including on TNT. %

\begin{table}[t]
    \caption{\textbf{Quantitative results on DTU~\cite{jensen2014large}.} We highlight the \textbf{best} result for each backbone method with and without the proposed sampler. $^\dagger$ Requires SfM points.}
    \label{tab:results_dtu}
    \resizebox{1\linewidth}{!}{%
        \setlength{\tabcolsep}{2pt}\begin{NiceTabular}{l l l c c c c c c c c c c c c c c c | c}[code-before =%
        \rectanglecolor{gray!20}{3-3}{3-19}%
        \rectanglecolor{gray!20}{5-3}{5-19}%
        \rectanglecolor{gray!20}{7-3}{7-19}%
        \rectanglecolor{gray!20}{9-3}{9-19}%
        \rectanglecolor{gray!20}{11-3}{11-19}%
        \rectanglecolor{gray!20}{13-3}{13-19}%
        \rectanglecolor{gray!20}{15-3}{15-19}%
        \rectanglecolor{gray!20}{17-3}{17-19}%
        \rectanglecolor{gray!20}{19-3}{19-19}%
        \rectanglecolor{gray!20}{21-3}{21-19}%
        ]
            \toprule
             & &  & 24 & 37 & 40 & 55 & 63 & 65 & 69 & 83 & 97 & 105 & 106 & 110 & 114 & 118 & 122 & \thead{Mean} \\\cmidrule(){4-19}
            \multirow{11}{*}{\rotatebox[origin=c]{90}{\thead{PSNR $\uparrow$}}}
            & \multirow{7}{*}{\rotatebox[origin=c]{90}{\thead{Unmasked}}}
            & NeRF~\cite{mildenhall2020nerf}         & 26.24 & 25.74 & 26.79 & 27.57 & 31.96 & 31.50 & 29.58 & 32.78 & 28.35 & 32.08 & 33.49 & 31.54 & 31.00 & 35.59 & 35.51 & 30.65 \\
            && VolSDF\cite{yariv2021volume}           & 26.28 & 25.61 & 26.55 & 26.76 & 31.57 & 31.50 & 29.38 & 33.23 & 28.03 & 32.13 & 33.16 & 31.49 & 30.33 & 34.90 & 34.75 & 30.38 \\
            && RegSDF$^\dagger$~\cite{Zhang_2022_CVPR}& 24.78 & 23.06 & 23.47 & 22.21 & 28.57 & 25.53 & 21.81 & 28.89 & 26.81 & 27.91 & 24.71 & 25.13 & 26.84 & 21.67 & 28.25 & 25.31 \\ \cmidrule(){3-19} %
            && NeuS~\cite{wang2021neus}        & 23.85 & 27.63 & 27.16 & 29.4  & 32.71 & 33.1  & 30.58 & 34.25 & 29.97 & 33.69 & 35.34 & 32.81 & 31.96 & 36.72 & 37    & 31.74 \\
             &  & NeuS + Ours                             & 28.28 & 28.1 & 28.16 & 24.71 & 33.1 & 33.97 & 29.59 & 33.25 & 30.35 & 33.61 & 35.66 & 32.97 & 32.29 & 37.15 & 35.56 & 31.78 \\ \cmidrule(){3-19}
             & & Neuralangelo~\cite{li2023neuralangelo} & 30.64 & 27.78 & 32.70 & \textbf{34.18} & 35.15 & \textbf{35.89} & 31.47 & \textbf{36.82} & 30.13 & \textbf{35.92} & 36.61 & 32.60 & 31.20 & \textbf{38.41} & \textbf{38.05} & 33.84\\
             &  & Neuralangelo + Ours                     & \textbf{33.73} & \textbf{30.36} & \textbf{33.55} & 34.06 & \textbf{35.22} & 34.64 & \textbf{32.49} & 33.2  & \textbf{31.93} & 34.17 & \textbf{37.64} & \textbf{35.3}  & \textbf{34.01} & 38.04 & 37.87 & \textbf{34.41} \\ \cmidrule(){2-19}

             & \multirow{4}{*}{\rotatebox[origin=c]{90}{\thead{Masked}}}
             & NeuS~\cite{wang2021neus}             & 28.93 & 28.29 & 27.53 & 30.57 & 36.48 & 36.48 & 31.83 & 40.59 & 31.26 & 37.19 & 36.87 & 33.9  & 32.65 & 39.63 & 40.88 & 34.21 \\
             & & NeuS + Ours                            & 28.98 & 29.22 & 28.66 & 25.22 & 37.24 & 38.73 & 30.77 & 42.47 & 32.34 & 37.5 & 37.49 & 34.34 & 33.11 & 40.84 & 38.45 & 34.36 \\ \cmidrule(){3-19}
             & & Neuralangelo~\cite{li2023neuralangelo} & \textbf{35.21} & 31.76 & 35.12 & 38.16 & 41.17 & 40.46 & 34.39 & 44.22 & 34.09 & 40.8  & 40.8  & 37.24 & 34.92 & 42.36 & 43.56 & 38.28 \\
             & & Neuralangelo + Ours                    & 35.13 & \textbf{32.86} & \textbf{35.2}  & \textbf{38.51} & \textbf{41.41} & \textbf{41}    & \textbf{34.51} & \textbf{44.93} & \textbf{35.64} & \textbf{41.13} & \textbf{40.95} & \textbf{37.78} & \textbf{35.26} & \textbf{43.3}  & \textbf{44.59} & \textbf{38.81} \\
\midrule
             
            \multicolumn{2}{c}{\multirow{9}{*}{\rotatebox[origin=c]{90}{\thead{Chamfer (mm) $\downarrow$}}}}
            & NeRF~\cite{mildenhall2020nerf}         & 1.90 & 1.60 & 1.85 & 0.58 & 2.28 & 1.27 & 1.47 & 1.67 & 2.05 & 1.07 & 0.88 & 2.53 & 1.06 & 1.15 & 0.96 & 1.49 \\
            & & VolSDF\cite{yariv2021volume}           & 1.14 & 1.26 & 0.81 & 0.49 & 1.25 & 0.70 & 0.72 & 1.29 & 1.18 & 0.70 & 0.66 & 1.08 & 0.42 & 0.61 & 0.55 & 0.86 \\
            & & HF-NeuS~\cite{wang2022hf}              & 0.76 & 1.32 & 0.70 & 0.39 & 1.06 & 0.63 & 0.63 & 1.15 & 1.12 & 0.80 & 0.52 & 1.22 & 0.33 & 0.49 & 0.50 & 0.77 \\
            & & RegSDF$^\dagger$~\cite{Zhang_2022_CVPR}          & 0.60 & 1.41 & 0.64 & 0.43 & 1.34 & 0.62 & 0.60 & 0.90 & 0.92 & 1.02 & 0.60 & 0.59 & 0.30 & 0.41 & 0.39 & 0.72 \\
             & & NeuralWarp~\cite{darmon2022improving}  & 0.49 & 0.71 & 0.38 & 0.38 & \textbf{0.79} & 0.81 & 0.82 & \textbf{1.20} & 1.06 & \textbf{0.68} & 0.66 & \textbf{0.74} & 0.41 & 0.63 & 0.51 & 0.68 \\ \cmidrule(){3-19}
             & & NeuS~\cite{wang2021neus}               & 0.77 & 0.78 & 5.82 & 0.50 & 1.39 & 1.76 & 1.06 & 4.01 & 1.47 & 0.77 & 0.64 & 1.29 & 0.34 & 0.56 & 0.53 & 1.30 \\
             & & NeuS + Ours                            & 1.08 & 0.74 & 1.27 & 2.43 & 1.05 & 1.05 & 1.66 & 1.32 & 2.1  & 0.79 & 0.6 & 1.07 & 0.32 & 0.4  & 2.08 & 1.2  \\ \cmidrule(){3-19}
             & & Neuralangelo~\cite{li2023neuralangelo} & \textbf{0.37} & 0.72 & 0.35 & 0.35 & 0.87 & \textbf{0.54} & \textbf{0.53} & 1.29 & 0.97 & 0.73 & \textbf{0.47} & \textbf{0.74} & 0.32 & 0.41 & 0.43 & 0.61 \\
             & & Neuralangelo + Ours                     &   0.39 & \textbf{0.68} & \textbf{0.32} & \textbf{0.33} &  0.87 & 0.58 & \textbf{0.53} & 1.3  & \textbf{0.93} & 0.70  & 0.5  & \textbf{0.74} & \textbf{0.31} & \textbf{0.37} & \textbf{0.38} &   \textbf{0.6 } \\
            
            \bottomrule%
        \end{NiceTabular}%
     }%
\end{table}

\paperpar{Quantitative results}
For image synthesis on the DTU and the TNT datasets, we assess the proposed method through two distinct analyses: evaluating the image quality and examining the quality in masked regions, given our focus on sampling areas with surfaces.
Results are shown in~\cref{tab:results_dtu}. We observe that adding the proposed sampling improves image quality, outperforming NeuS and Neuralangelo by a mean of $0.04dB$ and $0.57dB$ on PSNR, respectively. Moreover, when only regions of interest are evaluated, we observe a higher improvement over NeuS ($0.15dB$) and about the same gain for Neuralangelo. %
3D reconstruction performance on the DTU and TNT datasets is shown in~\cref{tab:results_dtu,tab:results_tnt}. We notice that even while relying on SfM priors, RegSDF achieves worse reconstruction results while we improve against the backbones. In particular, we achieve state-of-the-art performance with our sampling strategy coupled with Neuralangelo.

\paperpar{Ablation study}
We ablate our sampling strategy using DTU. We follow previous methods and use a subset of DTU ($24$, $65$, $97$, and $122$) to ablate the method. Results are shown in~\cref{tab:ablations}. 
The ablations \textbf{A1vsA2} show that surface losses substantially improve 3D reconstruction while slightly decreasing image quality.
Ablations \textbf{A2vsA3} assess the impact of view dependency, where focusing the sampling on occluded areas introduces uncertainty, diminishing both image quality and 3D reconstruction accuracy.
\textbf{A2vsA4} illustrates the impact of additional ray samples near the surface. These samples improve image quality but have minimal effect on 3D reconstruction.
Ablations \textbf{A2vsA5} examines a constant value for $s$ (logistic scale in \cref{eq:logistic_fuction}) in probability updates. 
When using a constant low-value for $s$, the 3D image space sampling produces high-variance samples, preventing refinement and introducing uncertainty in the model. 

\begin{table}[t]
    \begin{minipage}{0.37\linewidth}%
    \caption{\textbf{Quantative results for TNT:} Small improvement in 3D from a decrease in image quality.%
    \label{tab:results_tnt}%
    }%
    \resizebox{1\linewidth}{!}{%
    \setlength{\tabcolsep}{5.5pt}%
    \begin{NiceTabular}{@{}lcc}[code-before =%
        \rectanglecolor{Gray!20}{3-1}{3-8}%
        ]
        \toprule
                  &   \thead{PSNR $\uparrow$} &   \thead{F1 $\uparrow$} \\
        \midrule
        Neuralangelo &  \textbf{25.99} & 0.58 \\
        Neuralangelo~$+$~Ours &  25.65 &   \textbf{0.59} \\
        \bottomrule
    \end{NiceTabular}%
    }%
    \end{minipage}\hfill%
    \noindent\begin{minipage}{0.55\linewidth}%
    \caption{\textbf{Ablations:} L--surface loss weight; VD--view dependency; RU--uniform ray sampling; FS--fixed $s$ probability update.%
    \label{tab:ablations}
    }%
    \resizebox{1\linewidth}{!}{%
    \setlength{\tabcolsep}{5.5pt}%
    \begin{NiceTabular}{@{}ccccccc}[code-before =%
        \rectanglecolor{Gray!20}{3-1}{3-8}%
        \rectanglecolor{Gray!20}{5-1}{5-8}%
        ]
        \toprule
        {\thead{Ablation}} & {\thead{L}} & {\thead{VD}} & {\thead{RU}} & {\thead{FS}} & {\thead{PSNR $\uparrow$}} & {\thead{Chamfer $\downarrow$}} \\ \midrule
        \textbf{A1} & 0 & {\large\cmark} & {\large\xmark} & {\large\xmark} & 35.44 & 0.82 \\
        \textbf{A2} & 500 & {\large\cmark} & {\large\xmark} & {\large\xmark} & 35.24 & 0.57 \\
        \textbf{A3} & 500 & {\large\xmark} & {\large\xmark} & {\large\xmark} & 34.91 & 1.04 \\
        \textbf{A4} & 500 & {\large\cmark} & {\large\cmark} & {\large\xmark} & 34.82 & 0.58 \\
        \textbf{A5} & 500 & {\large\cmark} & {\large\xmark} & {\large\cmark} & 35.22 & 0.58 \\
        \bottomrule
    \end{NiceTabular}%
    }%
    \end{minipage}%
\end{table}

\paperpar{Computational overhead}
Evaluation overhead is insignificant. The 3D reconstruction does not change since grid points are directly evaluated in the implicit surface network. For image synthesis, only the $\lambda$-axis is sampled since $u$ and $v$ are known (all pixels in the image). The overhead during evaluation is negligible (average $+0.7\%$) when compared between Neuralangelo with and without our sampling in DTU.
Training overhead comes from interpolating the camera weights and interpolating and sampling each axis for all cameras in the batch, which amounts to an additional $10\%$ of training time with our implementation.

\section{Discussion}\label{sec:discussion}

In this work, we explore the scene's implicit surface representation to define a novel sampling strategy for training implicit neural surface fields more effectively. We propose a view-dependent and occlusion-aware 3D orthographic projection space, where we conditionally sample the image coordinates and depth for each camera. We utilize this depth to regularize the scene's surface during training. %
Our strategy can be coupled with typical image sampling on neural surface pipelines and does not depend on specific backbones. Experiments show that our proposed sampling strategy improves both image synthesis and 3D reconstruction, preserving the foreground details of the scene. 

Future work involves extending our approach to incorporate other sources of information, such as semantics or point clouds, to reduce in-ray uncertainties.
\section*{Acknowledgments}\label{sec:acknowledgments}

Pais and Piedade were partially supported by LARSyS, Portuguese ``Funda\c{c}\~{a}o para a Ci\^{e}ncia e a Tecnologia'' (FCT) funding (DOI: {\tt 10.54499/LA/P/0083/2020}, {\tt 10.54499/UIDP/50009/2020}, and {\tt 10.54499/UIDB/50009/2020}). Pais was also partially supported by FCT grant {\tt PD/BD/150630/2020}.
Miraldo was supported exclusively by Mitsubishi Electric Research Laboratories.%
{%
\small
\bibliographystyle{splncs04}
\bibliography{references}

}
}{}%
\ifthenelse{\boolean{suppmat}}
{%
\renewcommand\thefigure{\thesection.\arabic{figure}}
\renewcommand\thetable{\thesection.\arabic{table}}
\renewcommand{\theequation}{\thesection.\arabic{equation}}
\title{\titleTemplate\\%
    {\sc\large(Supplementary Materials)}} 
\titlerunning{\titleRunning}
\author{\authorsTemplate}
\authorrunning{\authorRunning}
\institute{\instituteTemplate}
\maketitle%
\vspace{12pt}%
{\it \noindent%
In this supplementary material, we provide more \underline{qualitative results} to showcase the effectiveness of our sampling strategy at effectively rendering scenes from novel views. Further, we show \underline{additional ablation studies}, to demonstrate the effect of the choice of different hyperparameters on our sampling strategy. Finally, we end this %
document by offering a \underline{formal mathematical underpinning} %
of our method. 
The following summarizes the supplementary materials:
\begin{itemize}
    \item Qualitative results;
    \item Additional ablation studies;
    \item Additional details of our method.
\end{itemize}
}
\appendix
\startcontents[suppMat]
\section{Qualitative Results}

We provide qualitative comparisons between our approach and competing baselines. Owing to the flexibility of our method, we can incorporate it into any neural implicit surface rendering system. Therefore, we denote competing baselines by their name, while our proposed approach is denoted by the baseline name~+~Ours. 
We show image synthesis and reconstruction results for the  Tanks and Temples (TNT) dataset in~\cref{subsec_supmat:additional_tnt}, and additional results for the BMVS~\cite{yao2020blendedmvs} and DTU~\cite{jensen2014large} datasets in~\cref{subsec_supmat:additional_bmvs,subsec:additional_dtu}, respectively.

In addition, we observed that NeuralWarp's framework for evaluating the 3D meshes tends to remove critical surfaces, creating an unfair scenario when the method does not exhibit boundary artifacts. Therefore, in this supplementary material, we also present a qualitative evaluation of the SDF output with and without the filtering process for the DTU dataset in~\cref{subsec_supmat:mesh_filtering_sup}.

\subsection{Qualitative Results on Tanks and Temples}
\label{subsec_supmat:additional_tnt}

In this subsection, we present qualitative rendering and 3D reconstruction results on the TNT dataset, comparing the impact of introducing our sampling strategy into Neuralangelo~\cite{li2023neuralangelo} versus the vanilla approach. Overall, as shown in ~\cref{fig:tnt_barn,fig:tnt_caterpillar}, we observe improved performance over the baseline by incorporating our method into it, with the less observed areas in the scene benefiting the most %
by incorporating the proposed sampling and surface reconstruction loss strategies (see ~\cref{fig:tnt_barn}).
However, as discussed in Sec. 5.2 in the main paper, for 
sequences that are less object-oriented %
and feature larger-scale and more complex environments
the improvement is less significant due to the nature of the sampling methodology, which focuses more %
on the object-centric regions in the scene. %

\begin{figure}[t]
    \centering
    \begin{subfigure}{0.47\linewidth}
        \ZoomPictureG{figs/sup_material/neur_tnt_barn_crop}{0.58}{.60}{}{3}{0.53}{.32}
    \caption{Neuralangelo}
    \end{subfigure}\hfill
    \begin{subfigure}{0.47\linewidth}
        \ZoomPictureG{figs/sup_material/neur_w_sampler_tnt_barn_crop}{0.58}{.60}{}{3}{0.53}{.32}
    \caption{Neuralangelo~+~Ours}
    \end{subfigure}
    \caption{\textbf{TNT Barn}: This figure shows an example of a rendered image and normal map for Neuralangelo and Neuralangelo with our proposed sampler. With the sampler, the sequence achieves a better 3D structure and greater details in regions where there is less reach of light, %
    whereas the original Neuralangelo has a spurious surface in those regions.}
    \label{fig:tnt_barn}
\end{figure}

\begin{figure}[t]
    \centering
    \begin{subfigure}{0.47\linewidth}
        \ZoomPictureH{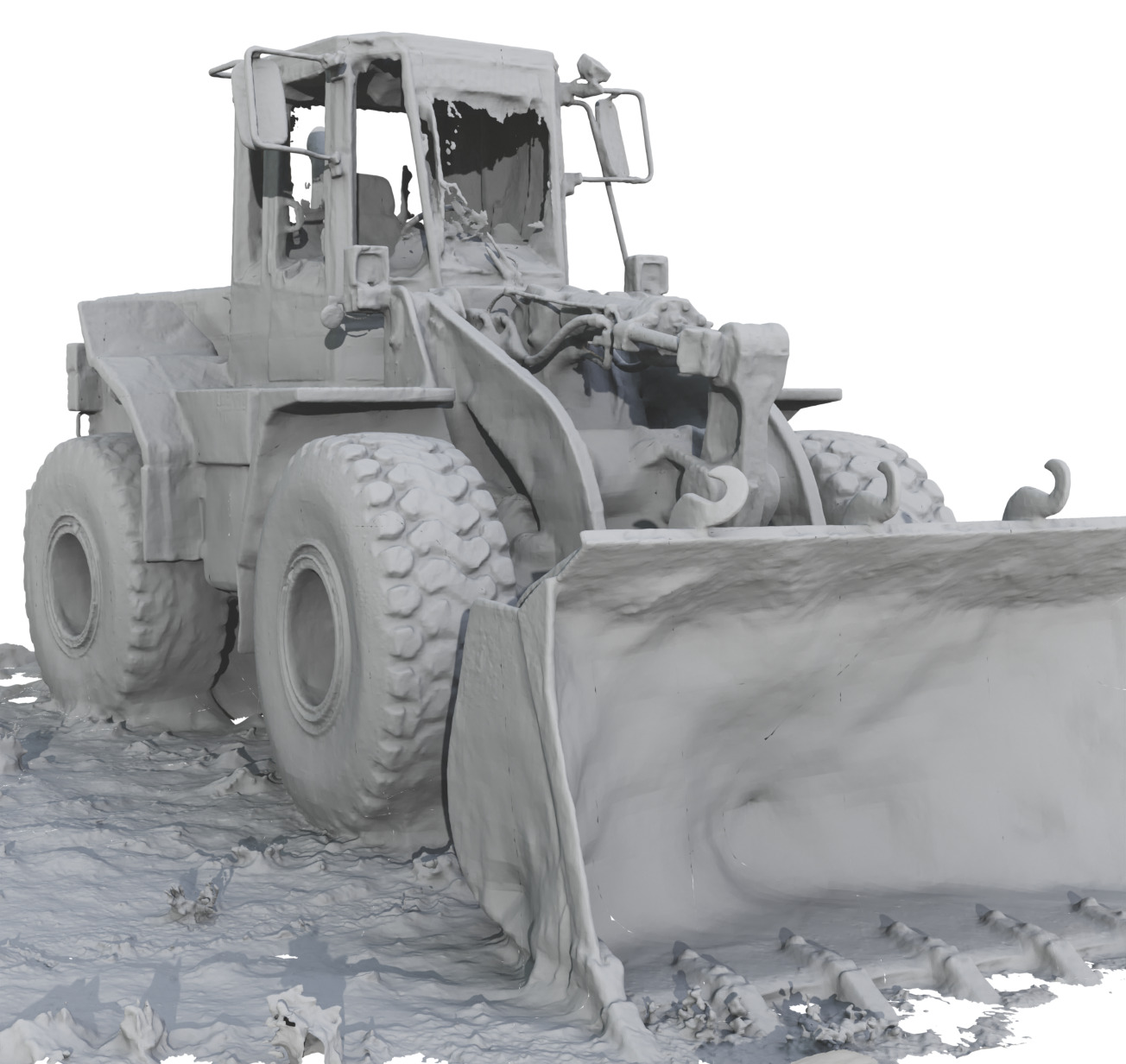}{0.87}{.07}{}{3}{0.48}{.68}
    \caption{Neuralangelo}
    \end{subfigure}\hfill
    \begin{subfigure}{0.47\linewidth}
        \ZoomPictureH{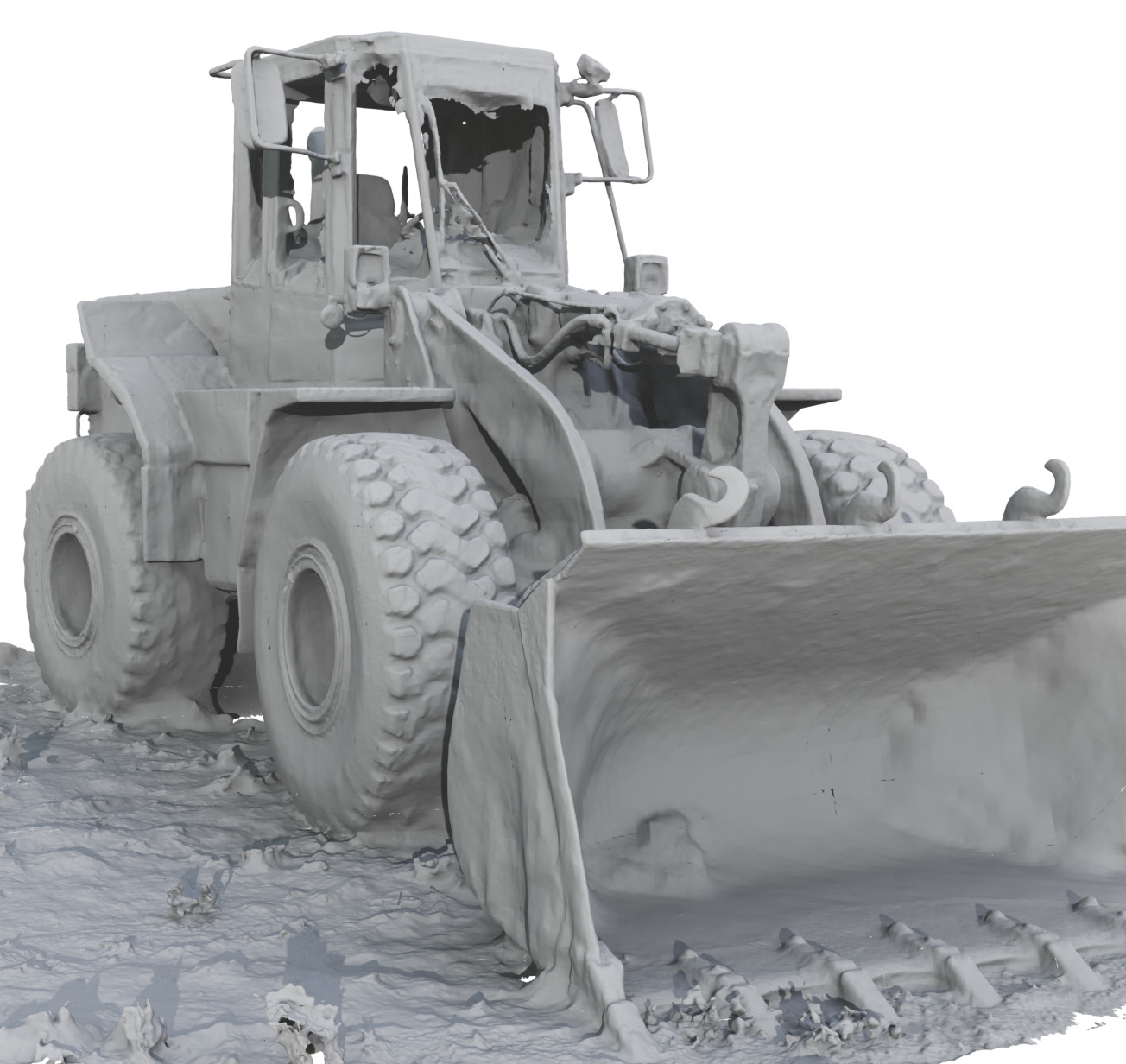}{0.87}{.07}{}{3}{0.48}{.68}
    \caption{Neuralangelo~+~Ours}
    \end{subfigure}
    \caption{\textbf{TNT Caterpillar}: In this figure, we show the 3D reconstruction of the Caterpillar sequence from TNT. The reconstruction from our methodology is more complete and has fewer unwanted surfaces, as shown in the caterpillar bucket. Additionally, there are more details and sharper surfaces, as seen on both wheels and highlighted regions.}
    \label{fig:tnt_caterpillar}
\end{figure}

\subsection{Qualitative Results on BMVS}
\label{subsec_supmat:additional_bmvs}
The BMVS~\cite{yao2020blendedmvs} dataset is a mixture of a large-scale environment, like TNT, and a more object-centric dataset, like DTU, where a bigger foreground/background separation exists. %
In~\cref{fig:bmvs_vase}, where the background is more complex, NeuS fails to reconstruct the 3D environment whereas NeuS+Ours performs well. We surmise that the absence of 3D consistency constraints, mainly affects image quality, given its role in retaining crucial information for preserving subtle surface textures directly associated with image sharpness. In contrast, the baselines are outperformed by incorporating the proposed sampler into them. NeuS with our sampler results are as good as Neuralangelo's. Neuralangelo with the proposed sampler shows better results, obtaining sharper details and better image quality.

\tikzset{
  path image/.style={
    path picture={
      \node at ([yshift=-0.1cm]path picture bounding box.center) {
        \includegraphics[height=2.7cm]{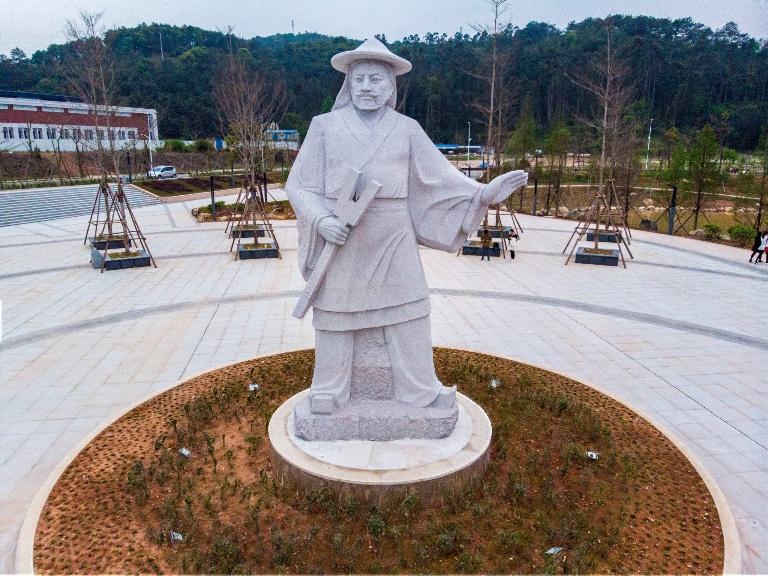}};}},
}

\begin{figure}[t]
    \centering
    \begin{subfigure}{0.47\textwidth}
        \ZoomPictureJ{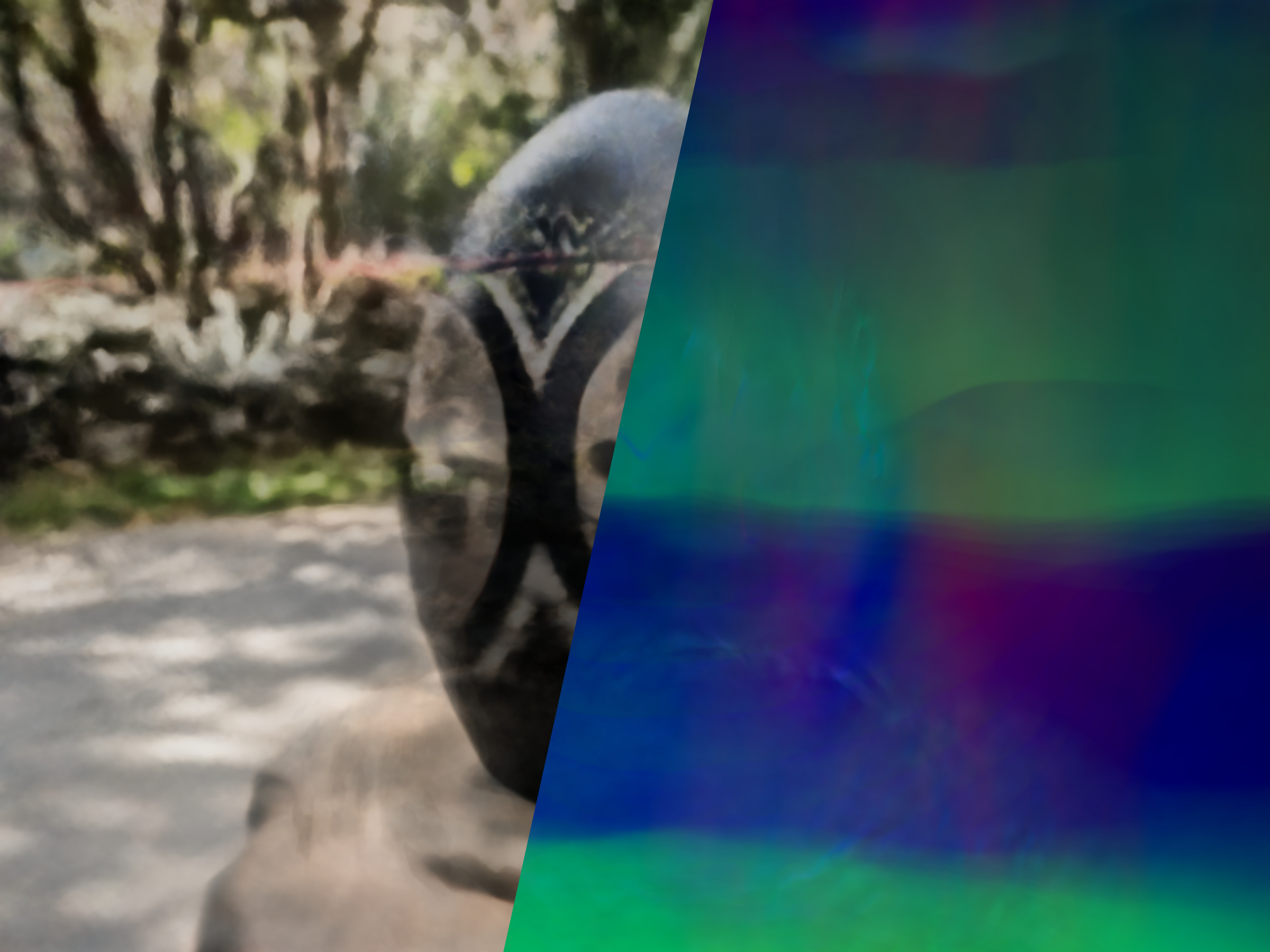}{0.51}{.39}{}{4}\vspace{10pt}
        \caption{NeuS}
    \end{subfigure}\hfill
    \begin{subfigure}{0.47\textwidth}
        \ZoomPictureJ{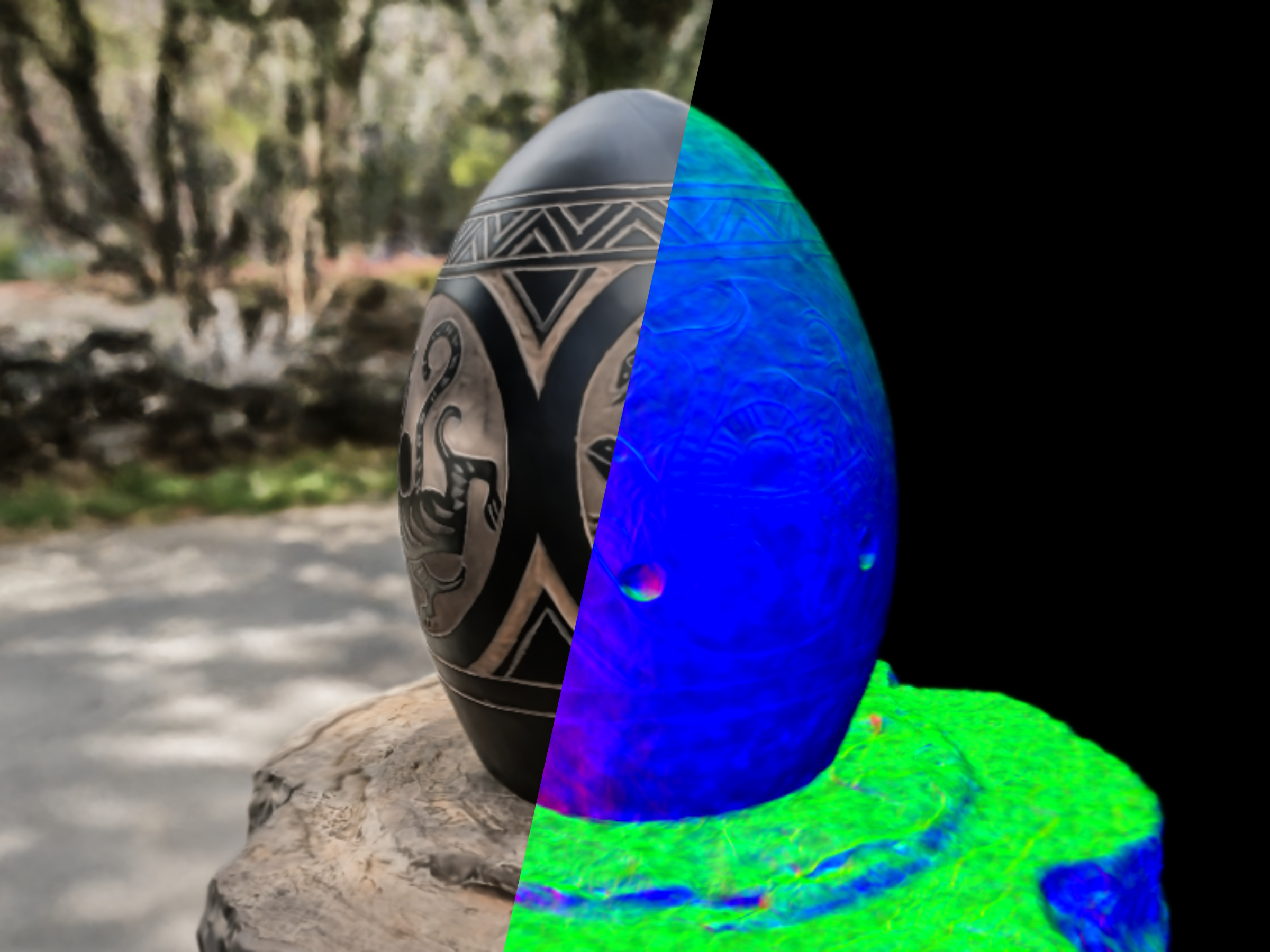}{0.51}{.39}{}{4}\vspace{10pt}
        \caption{NeuS~+~Ours}
    \end{subfigure}\\
    \begin{subfigure}{0.47\textwidth}
        \ZoomPictureJ{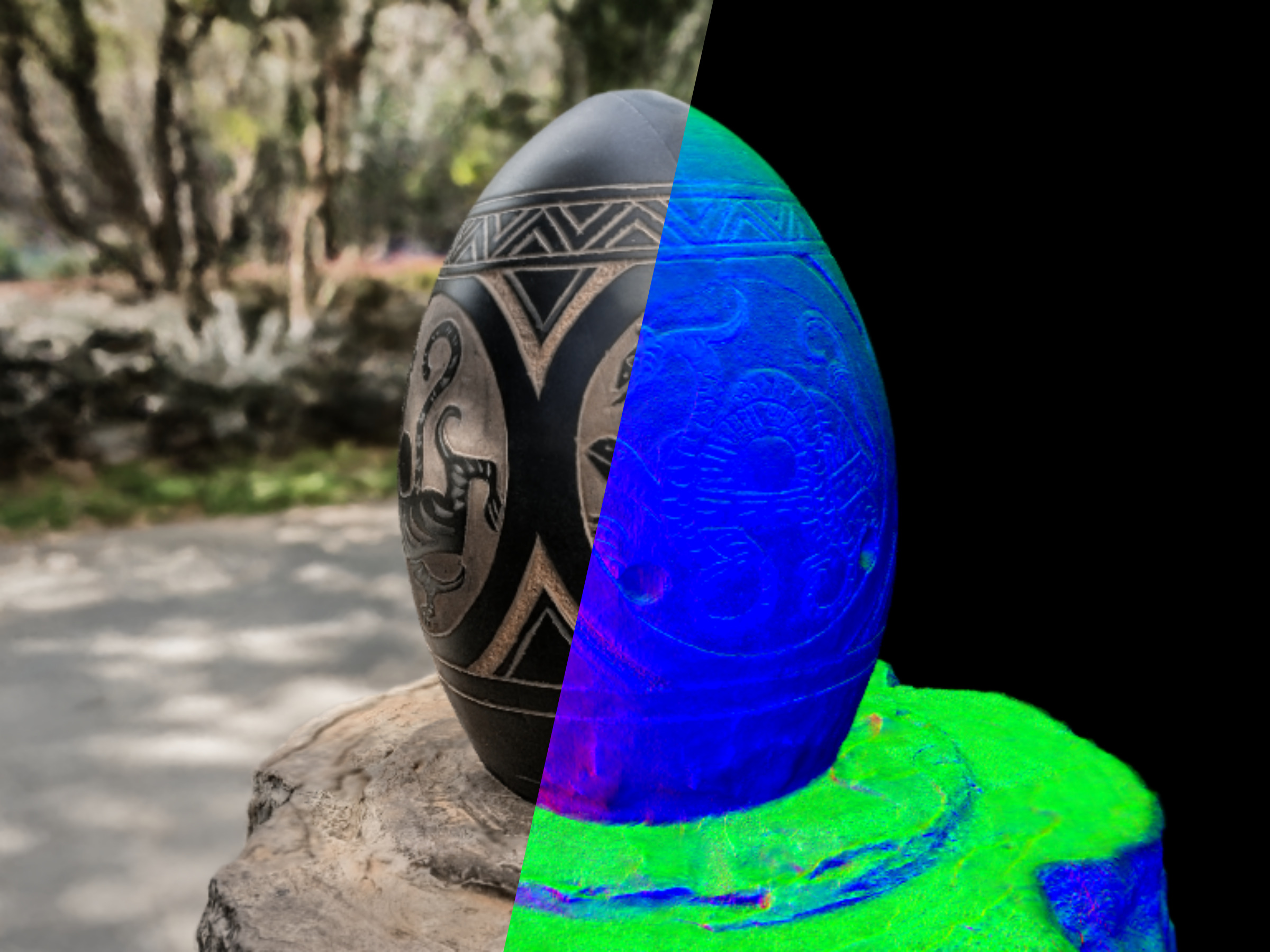}{0.51}{.39}{}{4}\vspace{10pt}
        \caption{Neuralangelo}
    \end{subfigure}\hfill
    \begin{subfigure}{0.47\textwidth}
        \ZoomPictureJ{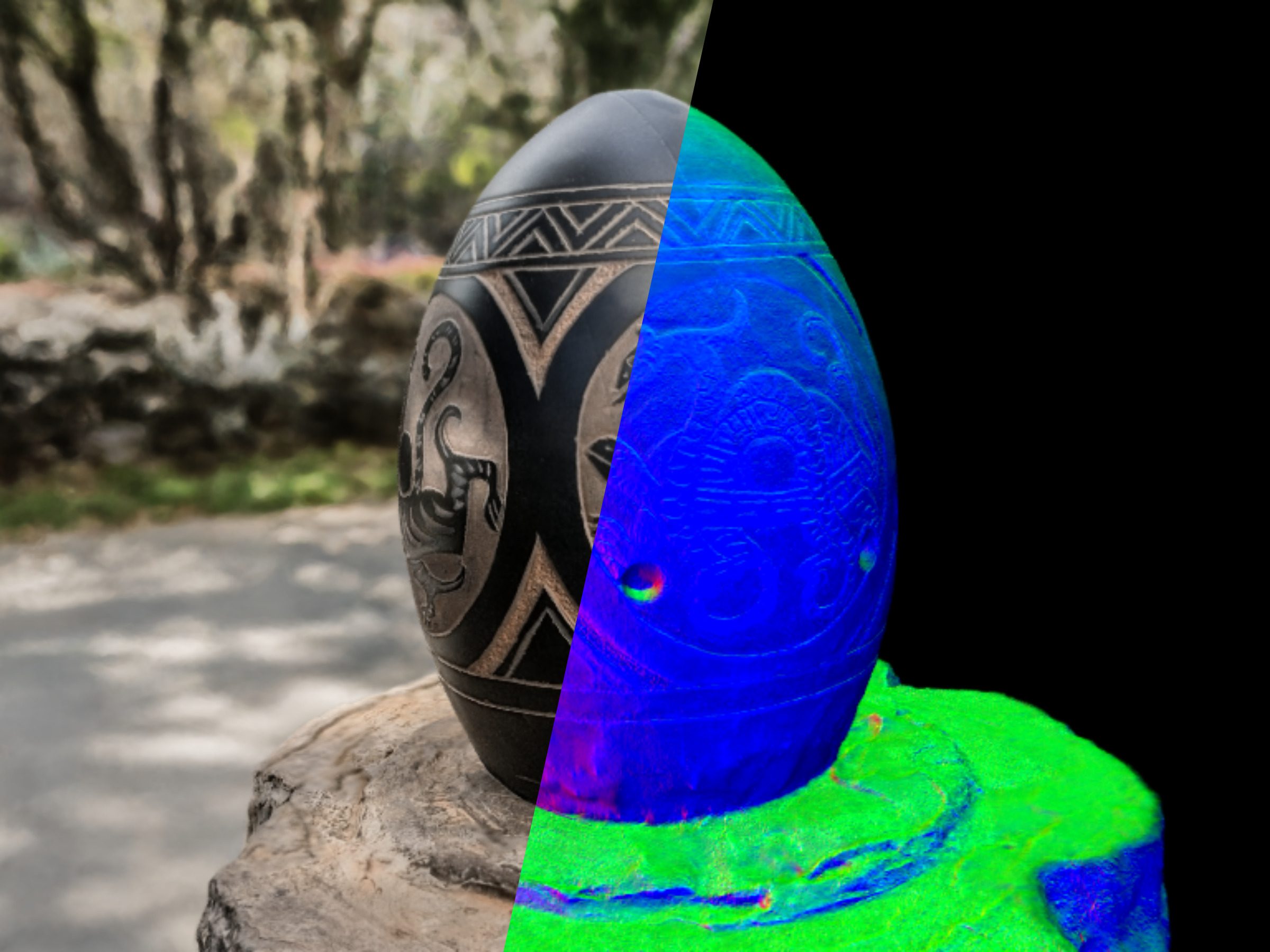}{0.51}{.39}{}{4}\vspace{10pt}
        \caption{Neuralangelo~+~Ours}
    \end{subfigure}%
    \caption{\textbf{BMVS "Vase"}: This figure shows the image and normal synthesis of a more complex scene in BMVS. NeuS is not able to reconstruct the scene. However, using our sampler with NeuS, we can reconstruct the scene. The same is true for Neuralangelo. The proposed methodology produces higher details and sharper edges.}
    \label{fig:bmvs_vase}
\end{figure}

\subsection{Qualitative Results on DTU}
\label{subsec:additional_dtu}

Upon further examination of the three datasets, the previously highlighted issues, pertaining to lack of 3D consistency, difficulty in rendering less observed portions of a scene, etc., persist when considering the DTU dataset. This is exemplified in~\cref{fig:dtu106,fig:dtu97}, by comparing the rendering results of Neuralangelo and NeuS with their counterparts that incorporate our sampling strategy. We hypothesize that the lack of detail in the 3D surfaces in the baselines is due to the absence of guided rays during training, which provides the necessary information to retain the details of the surface texture. This is especially noticeable in less frequently observed areas when rays are uniformly sampled in the pixel space, as shown in the images on the right (with our proposed sampling strategy) for each method in~\cref{fig:dtu97}. The additional samples and targeting of textured regions contribute to more detailed visual representations, as shown in~\cref{fig:dtu106}.

\begin{figure}[t]
    \centering
    \begin{subfigure}{0.48\linewidth}
        \ZoomPictureK{figs/sup_material/neus_dtu106_front}{0.57}{0.4}{}{2}\hspace{24\linewidth}
        \ZoomPictureL{figs/sup_material/neus_dtu106_back}{0.6}{.75}{}{4}\vspace{10pt}
        \caption{NeuS}
    \end{subfigure}\hfill
    \begin{subfigure}{0.48\linewidth}
        \ZoomPictureK{figs/sup_material/neus_w_sampler_dtu106_front}{0.57}{0.4}{}{2}\hspace{24\linewidth}
        \ZoomPictureL{figs/sup_material/neus_w_sampler_dtu106_back}{0.6}{.75}{}{4}\vspace{10pt}
        \caption{NeuS + Ours}
    \end{subfigure}\\
    \begin{subfigure}{0.47\columnwidth}
        \includegraphics[width=0.48\linewidth]{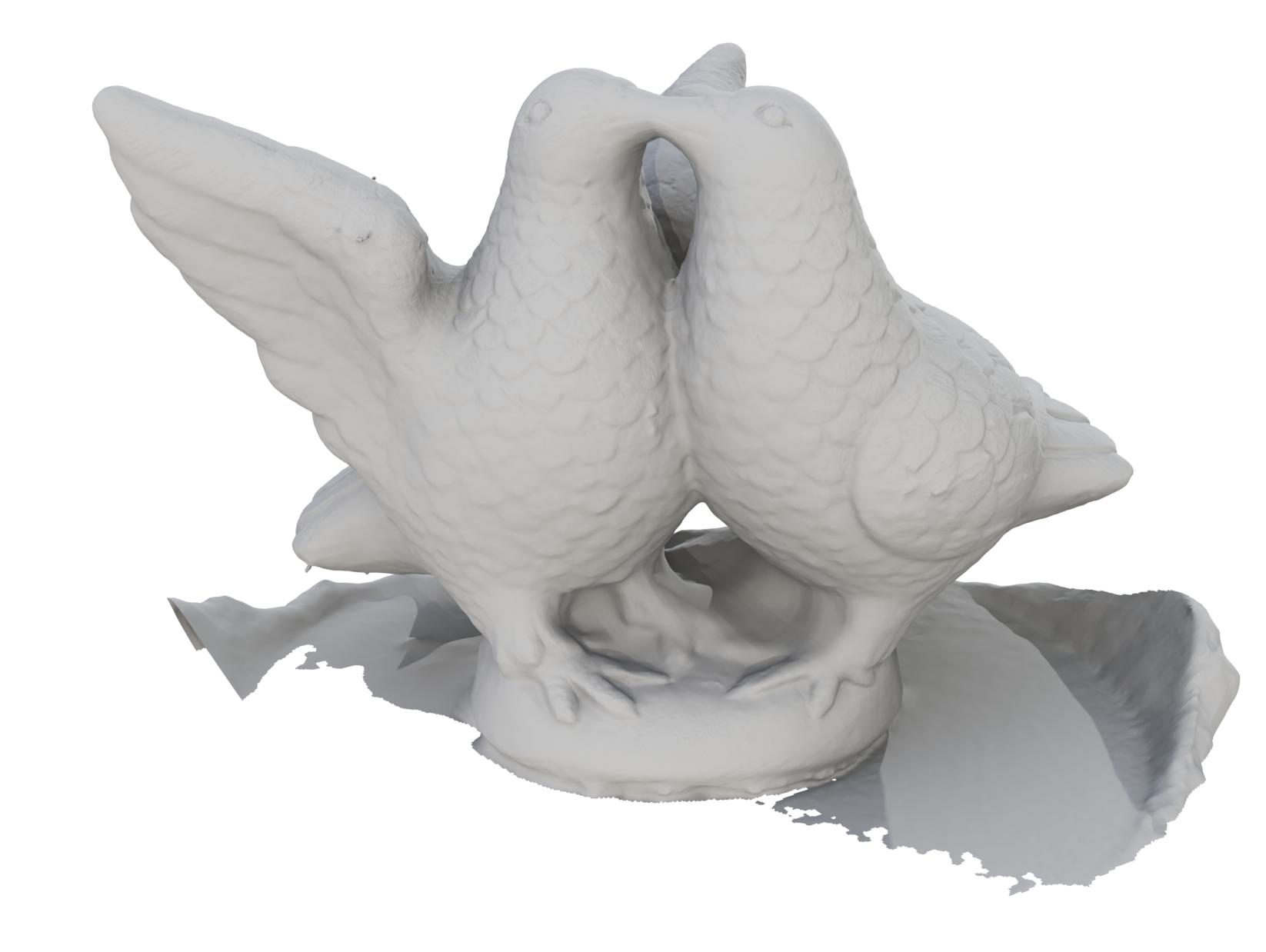}
\ZoomPictureL{figs/sup_material/neur_dtu106_back}{0.75}{.32}{}{2}\vspace{10pt}
        \caption{Neuralangelo}
    \end{subfigure}
    \begin{subfigure}{0.47\columnwidth}
        \includegraphics[width=0.48\linewidth]{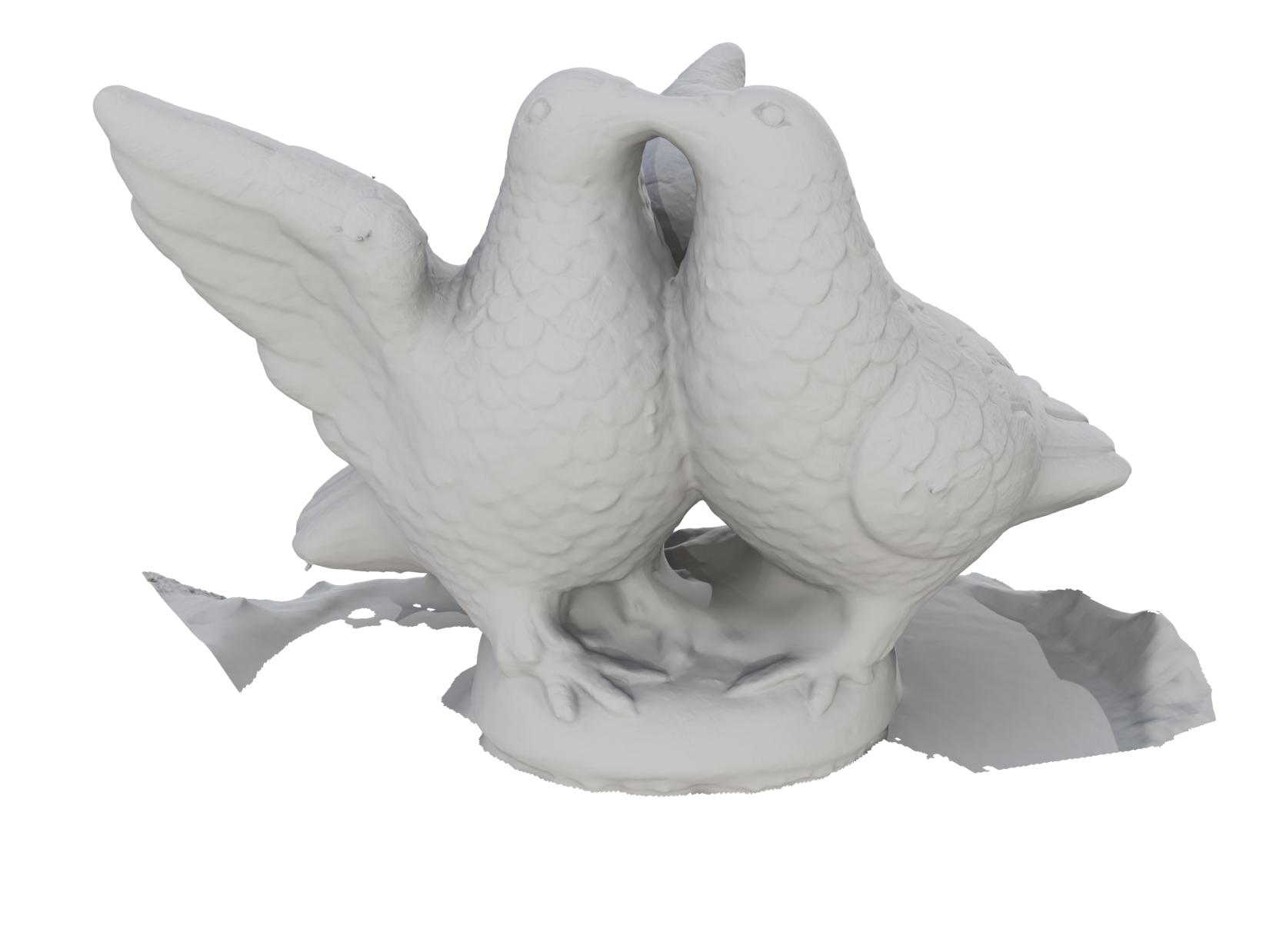}
\ZoomPictureL{figs/sup_material/neur_w_sampler_dtu106_back}{0.75}{.32}{}{2}\vspace{10pt}
        \caption{Neuralangelo + Ours}
    \end{subfigure}
    \caption{\textbf{DTU 106}: This figure shows the front (left) and back (right) reconstruction for a DTU scan. With our methodology, less observed regions are more complete for both backbones, with fewer unwanted surface areas in the back of each backbone.}%
    \label{fig:dtu106}
\end{figure}

\begin{figure}[t]
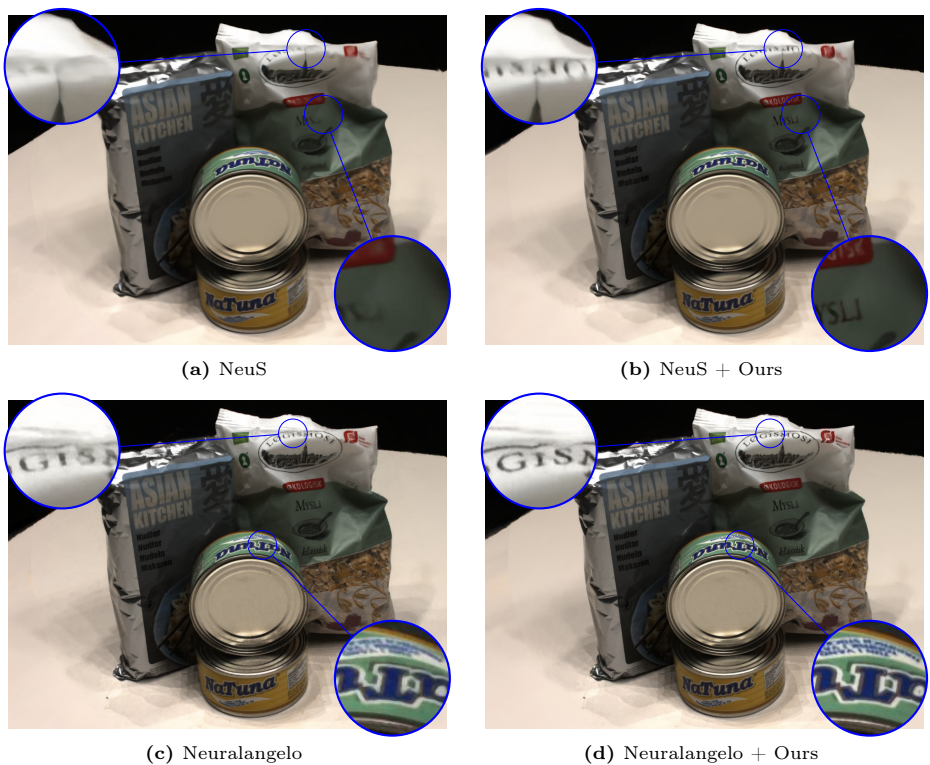

    \centering
    \begin{subfigure}{0.48\linewidth}
        \ZoomPictureF{figs/sup_material/neus_dtu97}{0.68}{.9}{}{3}{0.72}{.7}
        \caption{NeuS}
    \end{subfigure}\hfill
    \begin{subfigure}{0.48\linewidth}
        \ZoomPictureF{figs/sup_material/neus_w_sampler_dtu97}{0.68}{.9}{}{3}{0.72}{.7}
        \caption{NeuS~+~Ours}
    \end{subfigure}\\
    \begin{subfigure}{0.48\linewidth}
        \ZoomPictureF{figs/sup_material/neur_dtu97}{0.65}{.9}{}{4}{0.58}{.56}
        \caption{Neuralangelo}
    \end{subfigure}\hfill
    \begin{subfigure}{0.48\linewidth}
        \ZoomPictureF{figs/sup_material/neur_w_sampler_dtu97}{0.65}{.9}{}{4}{0.58}{.56}
        \caption{Neuralangelo~+~Ours}
    \end{subfigure}
    \caption{\textbf{DTU 97}: This figure shows texture synthesis using the proposed method. The additional samples and targeting of object regions contribute to more detailed visual representations, detailed by less blur and sharper details, compared with the respective backbones.} %
    \label{fig:dtu97}
\end{figure}

\subsection{Mesh Results without Filtering}
\label{subsec_supmat:mesh_filtering_sup}

\begin{figure}[t]
    \centering
    \begin{subfigure}{0.47\linewidth}
        \includegraphics[width=\textwidth]{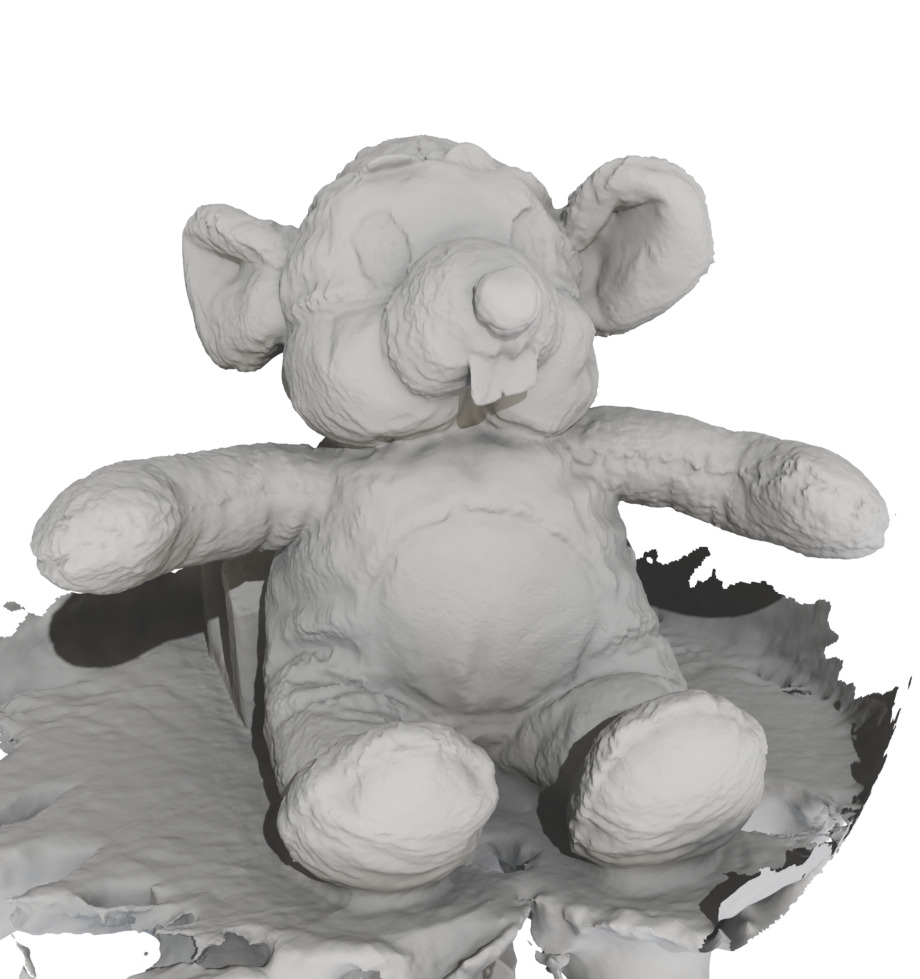}
    \caption{Without Filter}
    \end{subfigure}\hfill
    \begin{subfigure}{0.47\linewidth}
    \includegraphics[width=\textwidth]{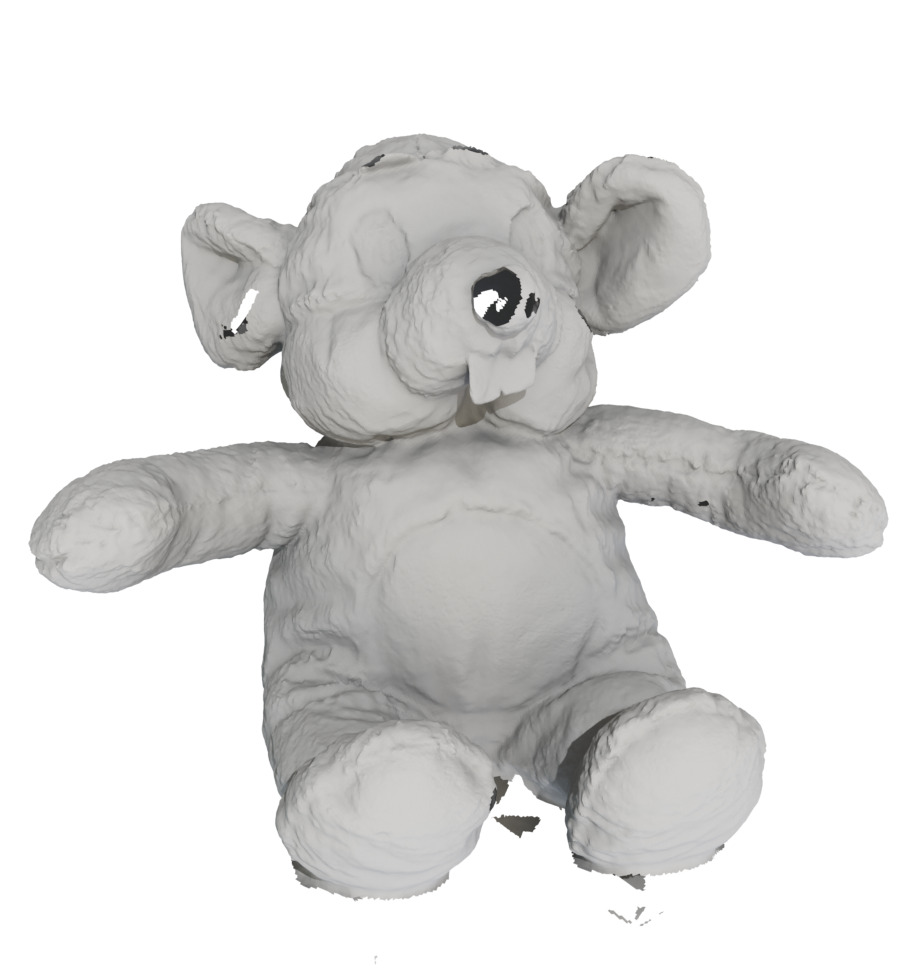}
    \caption{With Filter}
    \end{subfigure}
    \caption{\textbf{NeuralWarp's filtering}. NeuralWarp's camera filtering, which takes advantage of the masks, removes areas in front of the object (\eg, a hole in the nose) while removing the unmasked region.}
    \label{fig:neuralwarp_filter}
\end{figure}

\begin{figure}[t]
    \centering
    \begin{subfigure}{0.9\columnwidth}
        \includegraphics[width=0.475\linewidth]{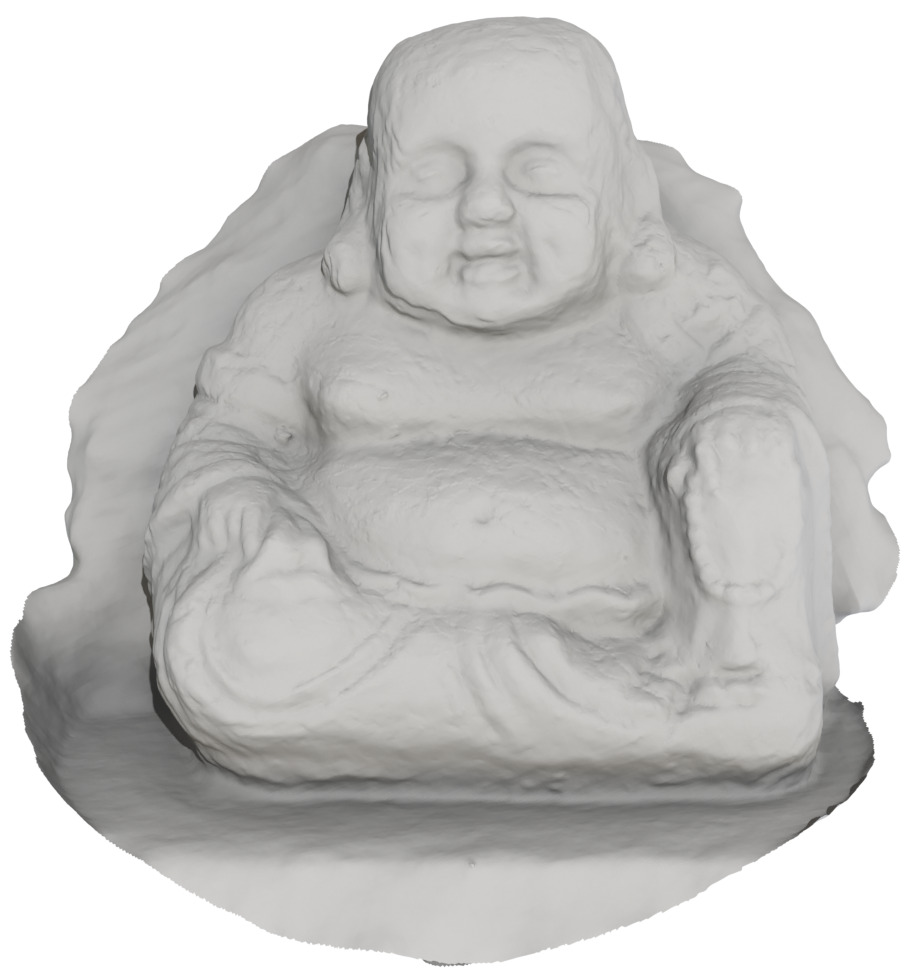}
        \includegraphics[width=0.475\linewidth]{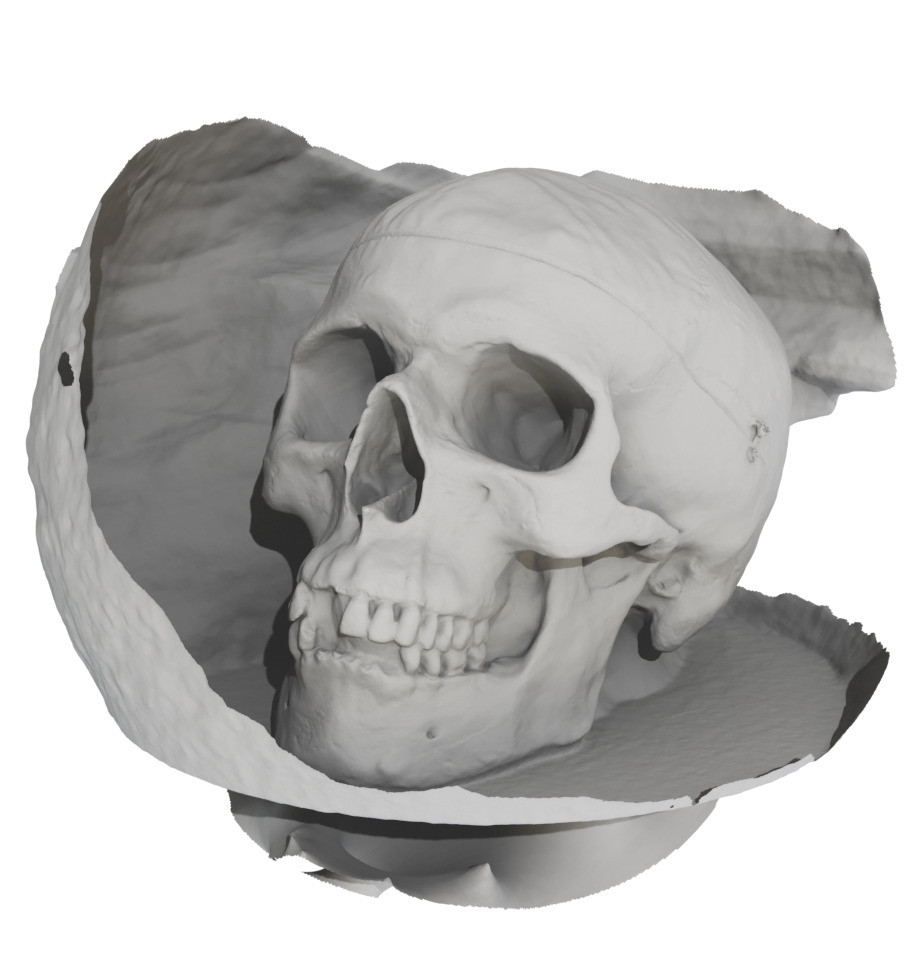}
        \caption{Neuralangelo}
    \end{subfigure}
    \\
    \begin{subfigure}{0.9\columnwidth}
        \includegraphics[width=0.475\linewidth]{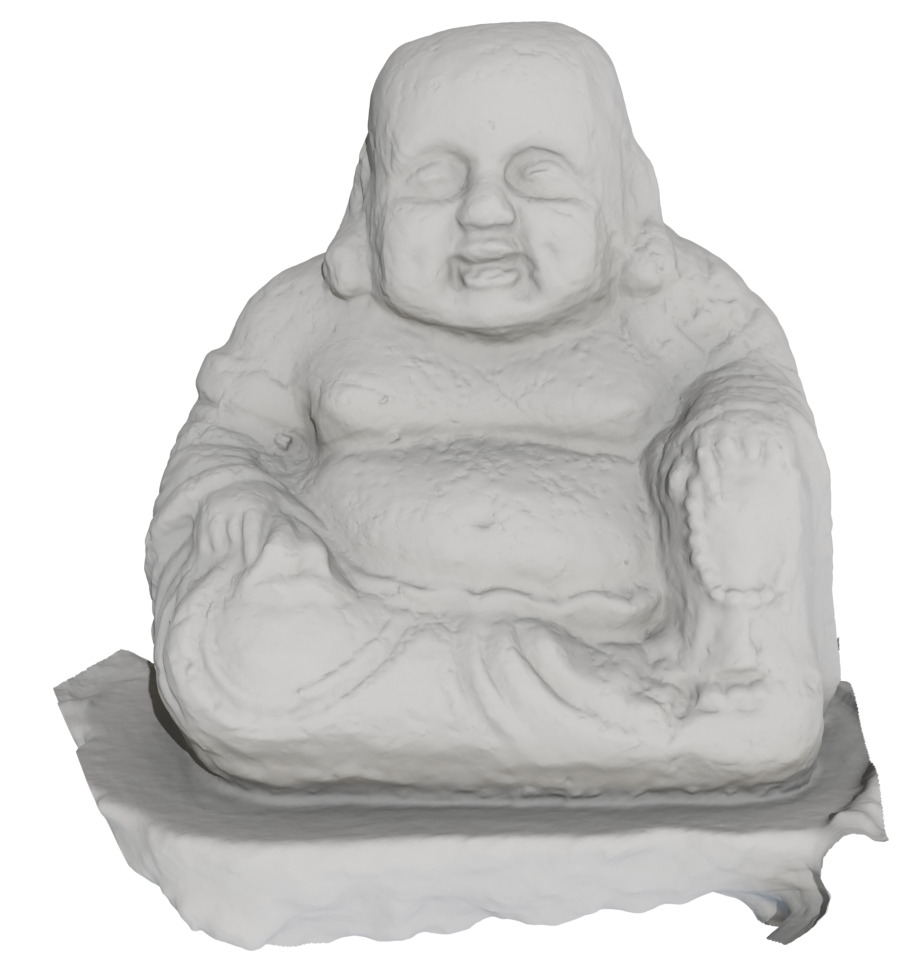}
        \includegraphics[width=0.475\linewidth]{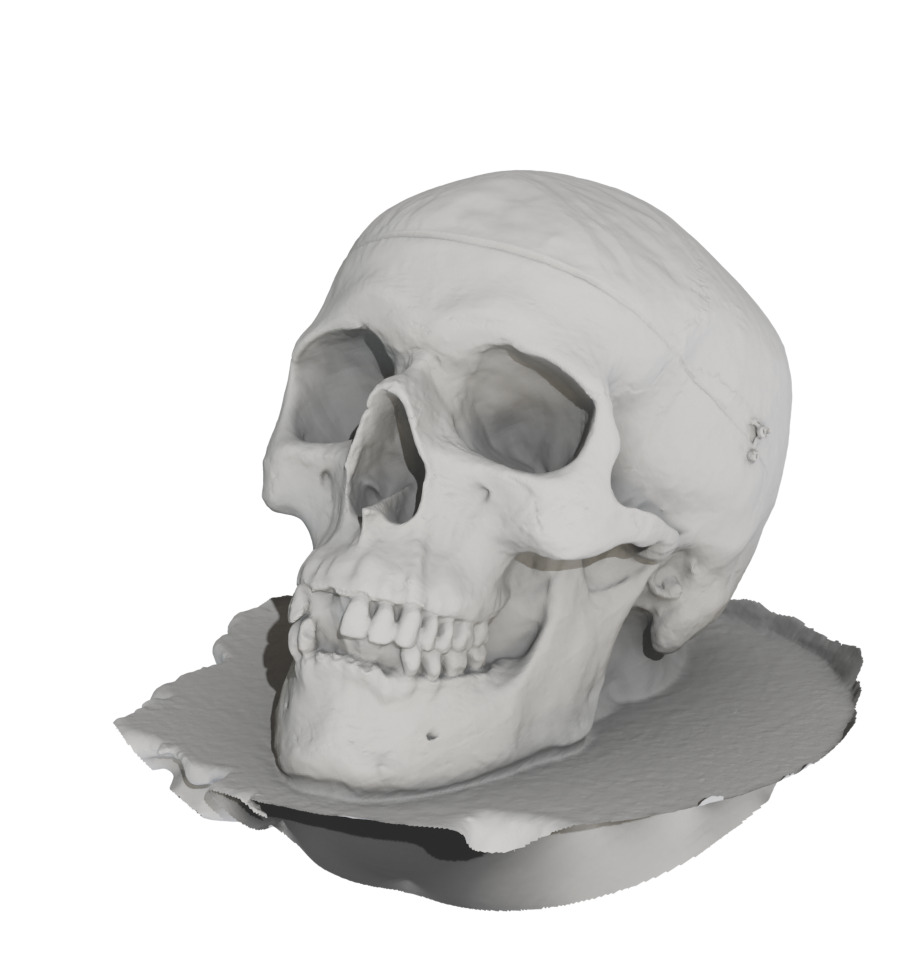}
        \caption{Neuralangelo~+~Ours}
    \end{subfigure}
    \caption{\textbf{Unfiltered SDF output}. This figure shows the SDF estimates of the output mesh without additional filtering. The proposed sampling with Neuralangelo reduces background boundary surfaces compared to the original Neuralangelo without any additional scene information.}
    \label{fig:without_filter}
\end{figure}

Analyzing the direct SDF output of Neuralangelo and Neuralangelo~+~Ours models, it is observed that the proposed sampling results in fewer boundary artifacts than the standard Neuralangelo, as shown in~\cref{fig:without_filter}. In scenes with more images where the floor is included, and there is no clear separation between the background, like the ones shown in~\cref{fig:without_filter}, Neuralangelo introduces surfaces around the boundary sphere on the back of the object. This becomes problematic when no additional information, such as masks or SfM points, is considered. Our proposed sampling, by directing rays towards regions with surfaces, effectively reduces the occurrence of spurious boundary surfaces.

NeuralWarp's framework\footnote{Framework used for evaluating the 3D meshes.} incorporates a camera-based filter to assess the models to eliminate such artifacts. However, this filtering approach tends to remove critical surfaces, which may be undesirable %
when the method does not exhibit boundary artifacts, as shown in~\cref{fig:neuralwarp_filter}. Nonetheless, consistent with prior work, the quantitative results presented in the paper use NeuralWarp's filter in the DTU dataset. Therefore, for a more holistic understanding of model performance, we show rendering results without this filtering in ~\cref{fig:neuralwarp_filter,fig:without_filter}.

\section{Additional Ablation Studies}\label{sec_supmat:ablations}

\begin{figure}[t]
    \begin{subfigure}[t]{1\linewidth}
        \includegraphics[width=1.0\linewidth]{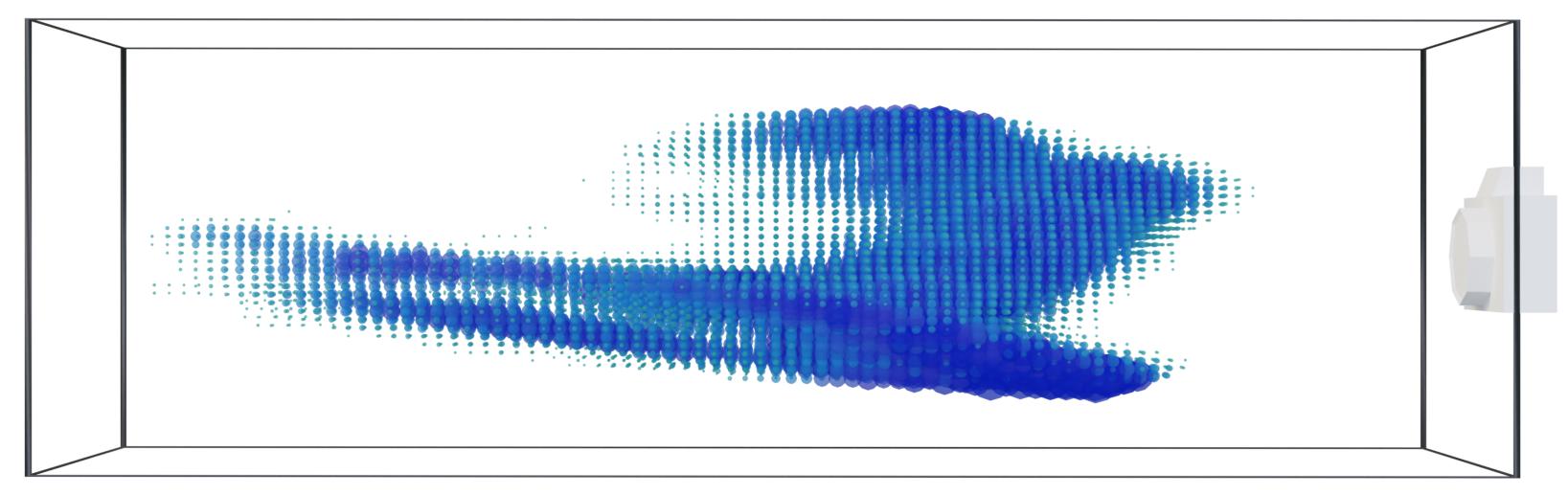}
        \caption{$F=1$ ($0.19s$)}
    \end{subfigure}
    \begin{subfigure}[t]{1\linewidth}
        \includegraphics[width=1.0\linewidth]{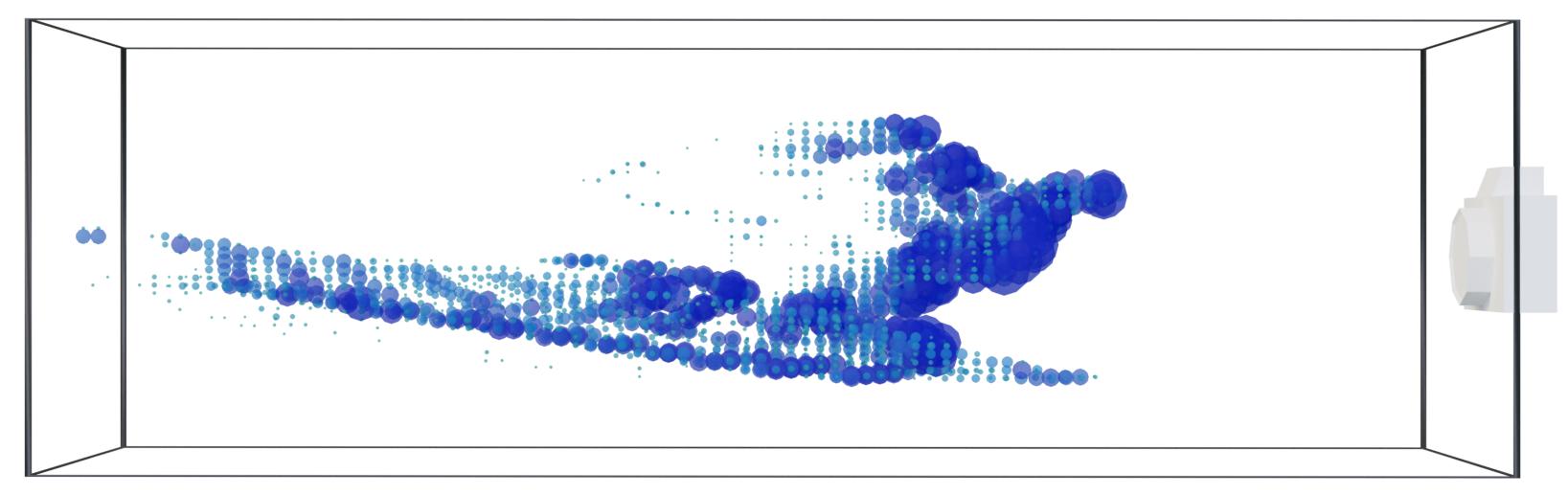}
        \caption{$F=2$ ($1.07s$)}
    \end{subfigure}
    \begin{subfigure}[t]{1\linewidth}\hfill
        \includegraphics[width=1.0\linewidth]{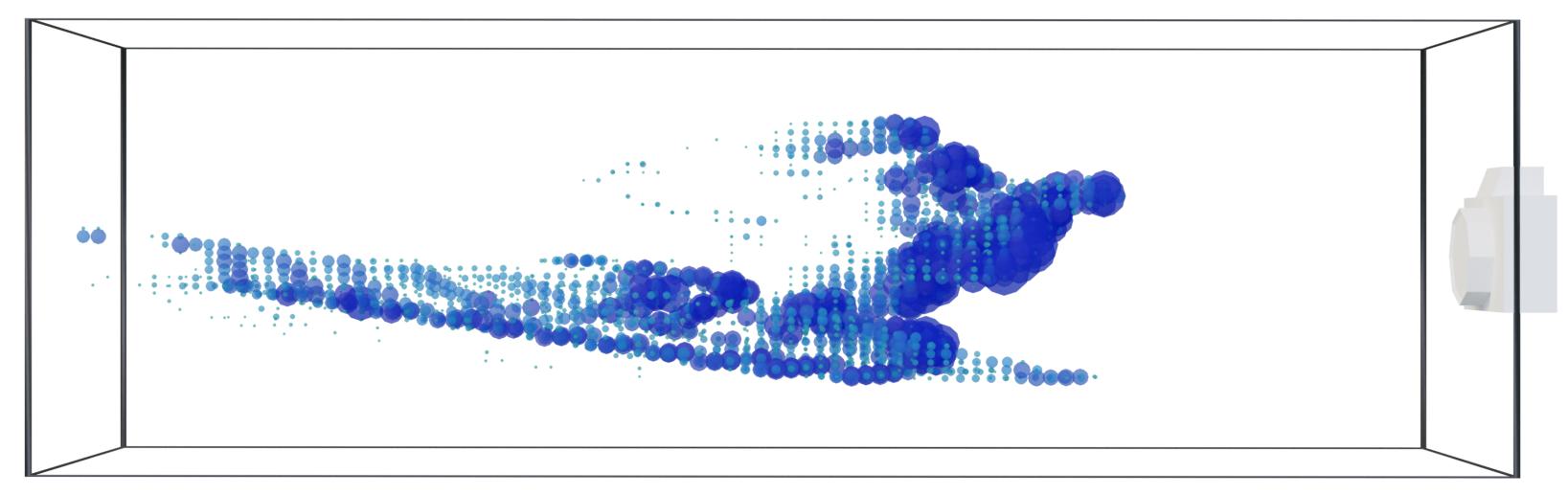}
        \caption{$F=4$ ($8.31s$)}
    \end{subfigure}
    \caption{\textbf{Interpolation factor}. This figure illustrates various interpolation factors for DTU scan $83$. The choice between $F = 2$ and $F > 2$ does not bring about a drastic alteration in normalized weights; however, raising the factor leads to a significant increase in update time. Decreasing it to $1$ results in a high variance PDF.}
    \label{fig:interpolation_factor}
\end{figure}

\begin{table}[t]
    \centering
    \caption{\textbf{Full Ablations:} L--surface reconstruction loss weight; F--partition factor; VD--view dependency; RU--uniform ray sampling; FS--fixed $s$ probability update; Log--using the logarithm of $s$ for ray sampling.}
    \label{tab:full_ablations}
    \resizebox{0.8\linewidth}{!}{%
    \setlength{\tabcolsep}{5.5pt}\begin{NiceTabular}{@{}ccccccccc}[code-before =%
        \rectanglecolor{Gray!20}{3-1}{3-9}%
        \rectanglecolor{Gray!20}{5-1}{5-9}%
        \rectanglecolor{Gray!20}{7-1}{7-9}%
        \rectanglecolor{Gray!20}{9-1}{9-9}%
        ]
        \toprule
        {\thead{Ablation}} & {\thead{L}} & {\thead{F}} & {\thead{VD}} & {\thead{RU}} & {\thead{FS}} & {\thead{Log}} & {\thead{PSNR $\uparrow$}} & {\thead{Chamfer $\downarrow$}} \\ \midrule
        \textbf{A0} & 0 & 1 & {\large\cmark} & {\large\xmark} & {\large\xmark} & {\large\xmark} & 34.71 & 0.87 \\
        \textbf{A1} & 0 & 2 & {\large\cmark} & {\large\xmark} & {\large\xmark} & {\large\xmark} & 35.44 & 0.82 \\
        \textbf{A2 - 10} & 10 & 2 & {\large\cmark} & {\large\xmark} & {\large\xmark} & {\large\xmark} & 35.44 & 0.85 \\
        \textbf{A2 - 500} & 500 & 2 & {\large\cmark} & {\large\xmark} & {\large\xmark} & {\large\xmark} & 35.24 & 0.57 \\
        \textbf{A3} & 500 & 2 & {\large\xmark} & {\large\xmark} & {\large\xmark}& {\large\xmark} & 34.91 & 1.04 \\
        \textbf{A4} & 500 & 2 & {\large\cmark} & {\large\cmark} & {\large\xmark} & {\large\xmark} & 34.82 & 0.58 \\
        \textbf{A5} & 500 & 2 & {\large\cmark} & {\large\xmark} & {\large\cmark} & {\large\xmark} & 35.22 & 0.57 \\
        \textbf{A6} & 500 & 2 & {\large\cmark} & {\large\xmark} & {\large\xmark} & {\large\cmark} & 35.22 & 0.58 \\
        \bottomrule
    \end{NiceTabular}}
\end{table}

In this section, we present additional ablation studies. For a further analysis of the proposed approach, see Sec. 5.2 of the main paper.

\paperpar{Interpolation}
As seen in~\cref{fig:interpolation_factor}, when interpolating the weights of the grids, there is a trade-off between accuracy and computational time controlled by the parameter $F$. The computational time is approximately proportional to $F^3$. Yet, little difference is seen in the weight interpolation when increasing from $F=2$ to $F=4$. Thus, we opted to let $F=2$ in the paper's experiments. Additionally, a higher $F$ reduces network uncertainty, as shown~\cref{tab:full_ablations}. Comparing \textbf{A0vsA1}, we observe that we improve image quality and 3D reconstruction by considering the additional partitions.

\paperpar{Different Weights for Surface Reconstruction Loss} From~\cref{tab:full_ablations}, we observe that increasing the weight for the surface reconstruction losses produces better overall results (\textbf{A1 vs A2 -- 10 vs A2 -- 500}). Modifying only the sampling scheme, \ie, guided sampling, without considering additional 3D information during training, does not substantially improve the 3D reconstruction results. Not using the loss or low values for the surface reconstruction loss, does not produce much higher qualitative and quantitative results. We observe the highest gains when the loss has a significant effect during training.

\paperpar{Ray Sampling Variance} As discussed in Sec. 4.5 of the main paper, additional samples are added to the ray, for improved performance. Evaluating \textbf{A2 -- 500 vs A4 and A6} shown in~\cref{tab:full_ablations}, we observe that the variance of the distribution from which these additional points are sampled does not make too much of a 
difference.
In \textbf{A6}, we sample ray points considering the Gaussian distribution $\sim \mathcal{N}(\tilde{\lambda}, \frac{\pi^2}{3 \log(s)^2})$, with higher variance than \textbf{A2}. Since, during the loss computation, $s$ is considered and not $\log(s)$, spreading more points along the ray has very little influence on the outcome.

\section{Additional Details of Our Method}

In this section, we provide more details about the proposed methodology. First, we offer a basic notation for the camera transformations and the probability densities in~\cref{supmat_subsec:notations}. Then, we present additional details for the implementation of the interpolation scheme in~\cref{supmat_subsec:interpolation}. 
Finally, we show how the ray is created from this space in~\cref{supmat_subsec:ray_sampling}.

\subsection{Notations and Camera Transformations}\label{supmat_subsec:notations}

This section defines the relevant notations, summarized in~\cref{tab:notations} and the additional details on the camera transformations.

\paperpar{Camera Transformations}
We define the transformations from the world to camera coordinates $c$, $h_c(\cdot): \mathbb{R}^3 \rightarrow \mathbb{R}^3$, and from the camera frame to the orthogonal projection space, $g(\cdot): \mathbb{R}^3 \rightarrow \mathbb{R}^3$. $h_c(\cdot)$ is defined by a rigid body transformation with rotation $\R_c \in \mathcal{SO}(3)$ and the camera location 
$\tvec_c \in\mathbb{R}^3$: 
\begin{equation}\label{eq:camera_transformation}
    \widehat \xvec_c = h_c(\xvec) = 
    \R^T_c \xvec -\R_c^T \tvec_c. 
\end{equation}
The projection $g(\cdot)$ is defined as
\begin{equation}
\uvec_c = g(\widehat \xvec_c) = 
\begin{bmatrix}
    \widehat x_c / \widehat z_c \\
    \widehat y_c / \widehat z_c \\
    \widehat z_c
\end{bmatrix} =
\begin{bmatrix}
    u_c \\
    v_c \\
    \lambda_c
\end{bmatrix}.
\end{equation}
The transformation from world to projection space is the composition of both functions: $f_c(\xvec) = g(h_c(\xvec))$.
In the same spirit as before, to simplify notation, we rewrite $\uvec = \uvec_c$, $\R = \R_c$, and $f(\cdot) = f_c(\cdot)$.

\begin{table}[t]
    \caption{Summary of some important notations.}
    \label{tab:notations}
    \small
    \centering
    \resizebox{0.85\linewidth}{!}{\setlength{\tabcolsep}{2.5pt}\begin{NiceTabular}{@{}cp{7.5cm}@{}}[code-before =%
        \rectanglecolor{Gray!20}{2-1}{2-2}%
        \rectanglecolor{Gray!20}{4-1}{4-2}%
        \rectanglecolor{Gray!20}{6-1}{6-2}%
        \rectanglecolor{Gray!20}{8-1}{8-2}%
        ]
    \toprule
        \thead{Notation} & \thead{Description}  \\
        \midrule
        $\xvec \in \mathcal{X}$  & A point in the 3D scene space. \\
        $\uvec \in \mathcal{U}$  & A point in the 3D orthographic projection space. \\
        $h(\cdot),\ g(\cdot),\ f(\cdot)$ & Camera extrinsic, intrinsic, and 3D projection. \\
        \multirow{2}{*}{$\vvec \in \mathcal{U}$}  & A point in the 3D orthogonal projection space from the transformation of $\xvec$, $f(\xvec)$. \\
        $\Gcal_{\Acal}$ & Grid in space $\Acal$. \\
        $p(\avec) $ & PDF defined by the weights and a grid $\Gcal_{\Acal}$.\\
        $S(\xvec)$ & SDF value for a 3D scene space point.\\
        $\phi_s(S(\xvec_i))$, $\sigma_i$, $\alpha_i$ & Probability value, density, and opacity at 3D point $i$. \\
        \bottomrule
    \end{NiceTabular}
    }
\end{table}

\subsection{Interpolation}\label{supmat_subsec:interpolation}

Given $p(\xvec)$ defined on a grid $\gridX$, we want to represent this PDF in camera coordinates over a grid $\gridU$. In this subsection, we aim to obtain this new PDF representing the probability distribution on the 3D image space $\Ucal$ (space of 3D parallel projection rays perpendicular to the image plane). To this end, we give a proposition that relates the weights in one grid to another via interpolation, accounting for the 3D projection $f(\cdot)$. The basic intuition is to account for the deformation of the cells through $f(\cdot)$, discretize the grid finely, and associate the probability mass with each smaller deformed cell to cells in the camera grid $\gridU$. Note that $\vvec$ is not represented in the grid of the three-dimensional image space $\gridU$. Instead, $\vvec$ is discretized according to the scene grid $\gridX$ after applying the camera transformation $f(\cdot)$, which we define as $\Gcal_{f(\Xcal)}$.

Using the change of coordinates transformation, we can equate $p(\xvec)$ such that 
\begin{equation}\label{eq:change_coord_sup2}
p(\xvec) = p(f^{-1}(\vvec))\Big|\frac{\partial f^{-1}(\vvec)}{\partial\vvec}\Big|.
\end{equation}
Now, applying the chain rule, one can write %
\begin{equation}\label{eq:partial_derivatives}
    \left| \frac{\partial f^{-1}(\vvec)}{\partial \vvec} \right| = \left| \frac{\partial h^{-1}(\hat{\xvec})}{\partial \hat\xvec} \right| \left| \frac{\partial g^{-1}(\vvec)}{\partial \vvec} \right|  = \left| \R \right| \left| \eta^2 \right| = \eta^2,
\end{equation}
where $\R$ is the rotation from the camera to world coordinates, defined in~\cref{eq:camera_transformation}. Note that in the coordinate system of $\vvec \in \Ucal$, $\eta$ is the distance orthogonal from the point to the image plane, \ie, the depth, which implies $\eta > 0$.
From~\cref{eq:partial_derivatives}, we can trivially prove that $h(\cdot)$ and $g(\cdot)$ are bijective since both have determinants different than zero and are invertible, by definition.
To compute the probability in the new coordinate system, we take \cref{eq:change_coord_sup2,eq:partial_derivatives} and solve for the probability in the image space
\begin{equation}\label{eq:change_var_sup}
    p(f^{-1}(\vvec)) = \eta^{-2} p(\xvec),  \ \ \forall \ \vvec \in f(\Xcal),
\end{equation}
considering the grids and their probabilities defined in Sec. 3.2, we take~\cref{eq:change_var_sup} and assume $\eta$ to be approximately constant in each cell.

We start by discretizing $\Gcal_{\Xcal}$ more finely to get the weights. Each cell with probability $p(\xvec)$ is partitioned into $F^3$ cells of equal volume with the probability of $p(\xvec_f)$, where $p(\xvec_f) = \frac{1}{F^3} p(\xvec)$ and $\xvec_f \in \Gcal_{\Xcal}^F \subset \Gcal_{\Xcal}$  . The partition divides each axis into $F$ equal parts. Then, we extend~\cref{eq:change_var_sup} for the partitioned cells in the transformed grid $\Gcal_{f(\Xcal)}^F$:
\begin{equation}\label{eq:change_var_part_sup}
    p(f^{-1}(\vvec_f)) = \eta^{-2} p(\xvec_f),  \ \ \forall \ \vvec_f \in \Gcal_{f(\Xcal)}^F,
\end{equation}
To interpolate the weights of $p(\uvec)$, we sum all $\vvec_f$ that are inside of the cell $\uvec$, as shown in~\cref{fig:interpolation_supmat}.
It is important to note that, with the increasing value of $F$, the volume of each partitioned cell converges to zero.

The computation of PDFs in the camera frame according to this interpolation scheme is appealing from a computational point of view, as the condition in the summation can be checked by $f(\xvec) \in \Ucal$ and vectorized. Furthermore, changing the parameter $F$ can control the computational complexity. We found it sufficient to let $F=2$, as motivated in~\cref{sec_supmat:ablations}. A representation of the interpolation scheme with the partitions is shown in~\cref{fig:interpolation_supmat}.

\begin{figure}
    \centering
    \scalebox{1.5}{\input{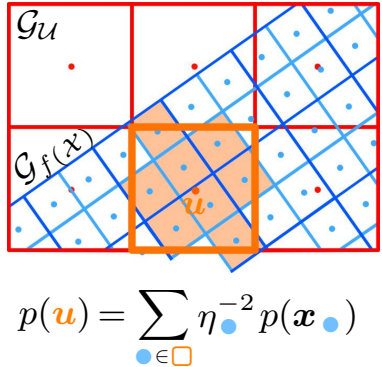}}
    \caption{\textbf{Illustration of the interpolation scheme considering the partitions.} This figure shows a representation of~\cref{eq:change_var_part_sup} and extends Fig. 3 of the main paper. First, the {\color{BlueGrid}scene grid} $\Gcal_{\Xcal}$ is partitioned into smaller cells with $F$, shown in {\color{BluePart}light blue}. Then, the partitioned grid is transformed according to $f(\cdot)$, $\Gcal_{f(\Xcal)}$, which intersects the {\color{red}image grid} $\Gcal_{\Ucal}$. Finally, to compute the weight on the image grid for each cell, the cell centers from the partitioned and transformed scene grid that lie within a specific {\color{BurntOrange}image space cell} are all summed according to their partitioned probability. Note that all smaller cells in the image space cell are evaluated in the inverse transformation to the scene space in the proposition. Still, the computation relates to the image space as depicted. 
    }
    \label{fig:interpolation_supmat}
\end{figure}

\paperpar{Relationship with Normalizing Flow}
\Cref{eq:change_var_sup} shows the change of variables of a probability distribution from calculus.
Normalizing flow is a generative technique to learn probability distributions from a sequence of latent space changes. Recent usage in neural fields~\cite{shen2022conditional} is to learn stochastic radiances and densities from the latent space and rays. The last latent spaces are the estimated radiance and the density used in the rendering.
Instead, our method approximates a camera density distribution from the surface representation and the world-to-image transformation. Our change of variables is fixed for each camera, and the probability is used for 3D image/ray sampling and not used in the rendering process. There are no stochastic sequences to learn from, as used in Normalizing flow methods.

\subsection{Ray Formation}\label{supmat_subsec:ray_sampling}

From the sampled point in the 3D image space $\widetilde \uvec$, we compute the ray for the neural rendering pipeline of the camera pose. We use the reverse transformation to the scene space to define the rays using $f^{-1}(\cdot)$ from the 3D image space.
The ray passing through the point $\widetilde \uvec$ is defined by a direction:
\begin{equation}
    \widetilde \dvec = \frac{f^{-1}(\widetilde \uvec) - \cvec}{||f^{-1}(\widetilde \uvec) - \cvec||},
\end{equation}
where $\cvec$ denotes the camera center and $||\cdot||$ is the euclidean norm. Subsequently, points in the world space along the ray are parameterized as $\xvec^{\widetilde \uvec} = \cvec + \lambda \widetilde \dvec$, where $\lambda > 0$ is the depth sampled along the ray.

\stopcontents[suppMat]%
\ifthenelse{\boolean{main}}
{}{%
\bibliographystyle{splncs04}%
\bibliography{references}%

\begin{thebibliography}{10}
\providecommand{\url}[1]{\texttt{#1}}
\providecommand{\urlprefix}{URL }
\providecommand{\doi}[1]{https://doi.org/#1}

\bibitem{attal2023hyperreel}
Attal, B., Huang, J.B., Richardt, C., Zollhoefer, M., Kopf, J., O’Toole, M., Kim, C.: Hyperreel: High-fidelity 6-dof video with ray-conditioned sampling. In: IEEE/CVF Conf. Computer Vision and Pattern Recognition (CVPR). pp. 16610--16620 (2023)

\bibitem{azinovic2022neural}
Azinovi{\'c}, D., Martin-Brualla, R., Goldman, D.B., Nie{\ss}ner, M., Thies, J.: Neural rgb-d surface reconstruction. In: IEEE/CVF Conf. Computer Vision and Pattern Recognition (CVPR). pp. 6290--6301 (2022)

\bibitem{barron2021mip}
Barron, J.T., Mildenhall, B., Tancik, M., Hedman, P., Martin-Brualla, R., Srinivasan, P.P.: Mip-nerf: A multiscale representation for anti-aliasing neural radiance fields. In: IEEE/CVF Int'l Conf. Computer Vision (ICCV). pp. 5855--5864 (2021)

\bibitem{Barron_2022_CVPR}
Barron, J.T., Mildenhall, B., Verbin, D., Srinivasan, P.P., Hedman, P.: Mip-nerf 360: Unbounded anti-aliased neural radiance fields. In: IEEE/CVF Conf. Computer Vision and Pattern Recognition (CVPR). pp. 5470--5479 (2022)

\bibitem{barron2023zip}
Barron, J.T., Mildenhall, B., Verbin, D., Srinivasan, P.P., Hedman, P.: Zip-nerf: Anti-aliased grid-based neural radiance fields. IEEE/CVF Conf. Computer Vision and Pattern Recognition (CVPR)  (2023)

\bibitem{campos2021orb}
Campos, C., Elvira, R., Rodr{\'\i}guez, J.J.G., Montiel, J.M., Tard{\'o}s, J.D.: Orb-slam3: An accurate open-source library for visual, visual--inertial, and multimap slam. IEEE Trans. Robotics (T-RO)  \textbf{37}(6),  1874--1890 (2021)

\bibitem{darmon2022improving}
Darmon, F., Bascle, B., Devaux, J.C., Monasse, P., Aubry, M.: Improving neural implicit surfaces geometry with patch warping. In: IEEE/CVF Conf. Computer Vision and Pattern Recognition (CVPR). pp. 6260--6269 (2022)

\bibitem{davison2003real}
Davison: Real-time simultaneous localisation and mapping with a single camera. In: IEEE Int'l Conf. Computer Vision (ICCV). pp. 1403--1410 (2003)

\bibitem{Deng_2023_ICCV}
Deng, J., Wu, Q., Chen, X., Xia, S., Sun, Z., Liu, G., Yu, W., Pei, L.: Nerf-loam: Neural implicit representation for large-scale incremental lidar odometry and mapping. In: IEEE/CVF Int'l Conf. Computer Vision (ICCV). pp. 8218--8227 (2023)

\bibitem{fu2022geo}
Fu, Q., Xu, Q., Ong, Y.S., Tao, W.: Geo-neus: Geometry-consistent neural implicit surfaces learning for multi-view reconstruction. Advances in Neural Information Processing Systems (NeurIPS)  \textbf{35},  3403--3416 (2022)

\bibitem{gaur2024orientedgrid}
Gaur, A., Pais, G.D., Miraldo, P.: Oriented-grid encoder for 3d implicit representations. In: Int'l Conf. 3D Vision (3DV) (2024)

\bibitem{geppert2020privacy}
Geppert, M., Larsson, V., Speciale, P., Sch{\"o}nberger, J.L., Pollefeys, M.: Privacy preserving structure-from-motion. In: European Conf. Computer Vision (ECCV). pp. 333--350 (2020)

\bibitem{Hartley2004}
Hartley, R.I., Zisserman, A.: Multiple View Geometry in Computer Vision. Cambridge University Press, 2 edn. (2004)

\bibitem{hu2023tri}
Hu, W., Wang, Y., Ma, L., Yang, B., Gao, L., Liu, X., Ma, Y.: Tri-miprf: Tri-mip representation for efficient anti-aliasing neural radiance fields. In: IEEE/CVF Int'l Conf. Computer Vision (ICCV). pp. 19774--19783 (2023)

\bibitem{jensen2014large}
Jensen, R., Dahl, A., Vogiatzis, G., Tola, E., Aan{\ae}s, H.: Large scale multi-view stereopsis evaluation. In: IEEE Conf. Computer Vision and Pattern Recognition (CVPR). pp. 406--413 (2014)

\bibitem{johnson2023unbiased}
Johnson, E., Habermann, M., Shimada, S., Golyanik, V., Theobalt, C.: Unbiased 4d: Monocular 4d reconstruction with a neural deformation model. In: IEEE/CVF Conf. Computer Vision and Pattern Recognition (CVPR). pp. 6597--6606 (2023)

\bibitem{Knapitsch2017}
Knapitsch, A., Park, J., Zhou, Q.Y., Koltun, V.: Tanks and temples: Benchmarking large-scale scene reconstruction. ACM Transactions on Graphics (TOG)  \textbf{36}(4) (2017)

\bibitem{Kong_2023_CVPR}
Kong, X., Liu, S., Taher, M., Davison, A.J.: vmap: Vectorised object mapping for neural field slam. In: IEEE/CVF Conf. Computer Vision and Pattern Recognition (CVPR). pp. 952--961 (2023)

\bibitem{kurz2022adanerf}
Kurz, A., Neff, T., Lv, Z., Zollh{\"o}fer, M., Steinberger, M.: Adanerf: Adaptive sampling for real-time rendering of neural radiance fields. In: European Conf. Computer Vision (ECCV). pp. 254--270 (2022)

\bibitem{Li_2022_CVPR}
Li, T., Slavcheva, M., Zollh\"ofer, M., Green, S., Lassner, C., Kim, C., Schmidt, T., Lovegrove, S., Goesele, M., Newcombe, R., Lv, Z.: Neural 3d video synthesis from multi-view video. In: IEEE/CVF Conf. Computer Vision and Pattern Recognition (CVPR). pp. 5521--5531 (2022)

\bibitem{li2023neuralangelo}
Li, Z., M{\"u}ller, T., Evans, A., Taylor, R.H., Unberath, M., Liu, M.Y., Lin, C.H.: Neuralangelo: High-fidelity neural surface reconstruction. In: IEEE/CVF Conf. Computer Vision and Pattern Recognition (CVPR). pp. 8456--8465 (2023)

\bibitem{Li_2021_CVPR}
Li, Z., Niklaus, S., Snavely, N., Wang, O.: Neural scene flow fields for space-time view synthesis of dynamic scenes. In: IEEE/CVF Conf. Computer Vision and Pattern Recognition (CVPR). pp. 6498--6508 (2021)

\bibitem{liu2020neural}
Liu, L., Gu, J., Zaw~Lin, K., Chua, T.S., Theobalt, C.: Neural sparse voxel fields. Advances in Neural Information Processing Systems (NeurIPS)  \textbf{33},  15651--15663 (2020)

\bibitem{liu2024gear}
Liu, X., Tai, Y.w., Tang, C.K., Miraldo, P., Lohit, S., Chatterjee, M.: Gear-nerf: Free-viewpoint rendering and tracking with motion-aware spatio-temporal sampling. In: IEEE/CVF Conf. Computer Vision and Pattern Recognition (CVPR) (2024)

\bibitem{mescheder2019occupancy}
Mescheder, L., Oechsle, M., Niemeyer, M., Nowozin, S., Geiger, A.: Occupancy networks: Learning 3d reconstruction in function space. In: IEEE/CVF Conf. Computer Vision and Pattern Recognition (CVPR). pp. 4460--4470 (2019)

\bibitem{mildenhall2020nerf}
Mildenhall, B., Srinivasan, P.P., Tancik, M., Barron, J.T., Ramamoorthi, R., Ng, R.: Nerf: Representing scenes as neural radiance fields for view synthesis. In: European Conf. Computer Vision (ECCV) (2020)

\bibitem{mosegaard1995monte}
Mosegaard, K., Tarantola, A.: Monte carlo sampling of solutions to inverse problems. Journal of Geophysical Research: Solid Earth  \textbf{100}(B7),  12431--12447 (1995)

\bibitem{muller2022instant}
M{\"u}ller, T., Evans, A., Schied, C., Keller, A.: Instant neural graphics primitives with a multiresolution hash encoding. ACM Transactions on Graphics (TOG)  \textbf{41}(4),  1--15 (2022)

\bibitem{Niemeyer_2020_CVPR}
Niemeyer, M., Mescheder, L., Oechsle, M., Geiger, A.: Differentiable volumetric rendering: Learning implicit 3d representations without 3d supervision. In: IEEE/CVF Conf. Computer Vision and Pattern Recognition (CVPR) (2020)

\bibitem{oechsle2021unisurf}
Oechsle, M., Peng, S., Geiger, A.: Unisurf: Unifying neural implicit surfaces and radiance fields for multi-view reconstruction. In: IEEE/CVF Int'l Conf. Computer Vision (ICCV). pp. 5589--5599 (2021)

\bibitem{Park_2021_ICCV}
Park, K., Sinha, U., Barron, J.T., Bouaziz, S., Goldman, D.B., Seitz, S.M., Martin-Brualla, R.: Nerfies: Deformable neural radiance fields. In: IEEE/CVF Int'l Conf. Computer Vision (ICCV). pp. 5865--5874 (2021)

\bibitem{peng2020convolutional}
Peng, S., Niemeyer, M., Mescheder, L., Pollefeys, M., Geiger, A.: Convolutional occupancy networks. In: European Conf. Computer Vision (ECCV). pp. 523--540 (2020)

\bibitem{pharr2023physically}
Pharr, M., Jakob, W., Humphreys, G.: Physically based rendering: From theory to implementation. MIT Press (2023)

\bibitem{Pumarola_2021_CVPR}
Pumarola, A., Corona, E., Pons-Moll, G., Moreno-Noguer, F.: D-nerf: Neural radiance fields for dynamic scenes. In: IEEE/CVF Conf. Computer Vision and Pattern Recognition (CVPR). pp. 10318--10327 (2021)

\bibitem{saito2020pifuhd}
Saito, S., Simon, T., Saragih, J., Joo, H.: Pifuhd: Multi-level pixel-aligned implicit function for high-resolution 3d human digitization. In: IEEE/CVF Conf. Computer Vision and Pattern Recognition (CVPR). pp. 84--93 (2020)

\bibitem{Sandstrom_2023_ICCV}
Sandstr\"om, E., Li, Y., Van~Gool, L., Oswald, M.R.: Point-slam: Dense neural point cloud-based slam. In: IEEE/CVF Int'l Conf. Computer Vision (ICCV). pp. 18433--18444 (2023)

\bibitem{Schonberger_2016_CVPR}
Schonberger, J.L., Frahm, J.M.: Structure-from-motion revisited. In: IEEE Conf. Computer Vision and Pattern Recognition (CVPR) (2016)

\bibitem{schonberger2016pixelwise}
Sch{\"o}nberger, J.L., Zheng, E., Frahm, J.M., Pollefeys, M.: Pixelwise view selection for unstructured multi-view stereo. In: European Conf. Computer Vision (ECCV). pp. 501--518 (2016)

\bibitem{shen2022conditional}
Shen, J., Agudo, A., Moreno-Noguer, F., Ruiz, A.: Conditional-flow nerf: Accurate 3d modelling with reliable uncertainty quantification. In: European Conf. Computer Vision (ECCV). pp. 540--557 (2022)

\bibitem{stefanski1991normal}
Stefanski, L.A.: A normal scale mixture representation of the logistic distribution. Statistics \& Probability Letters  \textbf{11}(1),  69--70 (1991)

\bibitem{Sucar_2021_ICCV}
Sucar, E., Liu, S., Ortiz, J., Davison, A.J.: imap: Implicit mapping and positioning in real-time. In: IEEE/CVF Int'l Conf. Computer Vision (ICCV). pp. 6229--6238 (2021)

\bibitem{sun2022direct}
Sun, C., Sun, M., Chen, H.T.: Direct voxel grid optimization: Super-fast convergence for radiance fields reconstruction. In: IEEE/CVF Conf. Computer Vision and Pattern Recognition (CVPR). pp. 5459--5469 (2022)

\bibitem{sun2022neural}
Sun, J., Chen, X., Wang, Q., Li, Z., Averbuch-Elor, H., Zhou, X., Snavely, N.: Neural 3d reconstruction in the wild. In: ACM SIGGRAPH (2022)

\bibitem{sun2021neuralrecon}
Sun, J., Xie, Y., Chen, L., Zhou, X., Bao, H.: Neuralrecon: Real-time coherent 3d reconstruction from monocular video. In: IEEE/CVF Conf. Computer Vision and Pattern Recognition (CVPR). pp. 15598--15607 (2021)

\bibitem{sun2024efficient}
Sun, S., Liu, M., Fan, Z., Jiao, Q., Liu, Y., Dong, L., Kong, L.: Efficient ray sampling for radiance fields reconstruction. Computers \& Graphics  \textbf{118},  48--59 (2024)

\bibitem{Tancik_2022_CVPR}
Tancik, M., Casser, V., Yan, X., Pradhan, S., Mildenhall, B., Srinivasan, P.P., Barron, J.T., Kretzschmar, H.: Block-nerf: Scalable large scene neural view synthesis. In: IEEE/CVF Conf. Computer Vision and Pattern Recognition (CVPR). pp. 8248--8258 (2022)

\bibitem{Tretschk_2021_ICCV}
Tretschk, E., Tewari, A., Golyanik, V., Zollh\"ofer, M., Lassner, C., Theobalt, C.: Non-rigid neural radiance fields: Reconstruction and novel view synthesis of a dynamic scene from monocular video. In: IEEE/CVF Int'l Conf. Computer Vision (ICCV). pp. 12959--12970 (2021)

\bibitem{turki2022mega}
Turki, H., Ramanan, D., Satyanarayanan, M.: Mega-nerf: Scalable construction of large-scale nerfs for virtual fly-throughs. In: IEEE/CVF Conf. Computer Vision and Pattern Recognition (CVPR). pp. 12922--12931 (2022)

\bibitem{wang2022neuris}
Wang, J., Wang, P., Long, X., Theobalt, C., Komura, T., Liu, L., Wang, W.: Neuris: Neural reconstruction of indoor scenes using normal priors. In: European Conf. Computer Vision (ECCV). pp. 139--155 (2022)

\bibitem{wang2021neus}
Wang, P., Liu, L., Liu, Y., Theobalt, C., Komura, T., Wang, W.: Neus: Learning neural implicit surfaces by volume rendering for multi-view reconstruction. In: Advances in Neural Information Processing Systems (NeurIPS) (2021)

\bibitem{wang2022neus2}
Wang, Y., Han, Q., Habermann, M., Daniilidis, K., Theobalt, C., Liu, L.: Neus2: Fast learning of neural implicit surfaces for multi-view reconstruction. In: IEEE/CVF Int'l Conf. Computer Vision (ICCV). pp. 3295--3306 (2023)

\bibitem{wang2022hf}
Wang, Y., Skorokhodov, I., Wonka, P.: Hf-neus: Improved surface reconstruction using high-frequency details. Advances in Neural Information Processing Systems (NeurIPS)  \textbf{35},  1966--1978 (2022)

\bibitem{wei2020deepsfm}
Wei, X., Zhang, Y., Li, Z., Fu, Y., Xue, X.: Deepsfm: Structure from motion via deep bundle adjustment. In: European Conf. Computer Vision (ECCV). pp. 230--247 (2020)

\bibitem{wen2023bundlesdf}
Wen, B., Tremblay, J., Blukis, V., Tyree, S., M{\"u}ller, T., Evans, A., Fox, D., Kautz, J., Birchfield, S.: Bundlesdf: Neural 6-dof tracking and 3d reconstruction of unknown objects. In: IEEE/CVF Conf. Computer Vision and Pattern Recognition (CVPR). pp. 606--617 (2023)

\bibitem{wu2023actray}
Wu, J., Liu, L., Tan, Y., Jia, Q., Zhang, H., Zhang, X.: Actray: Online active ray sampling for radiance fields. In: ACM SIGGRAPH Asia. pp. 1--10 (2023)

\bibitem{yao2020blendedmvs}
Yao, Y., Luo, Z., Li, S., Zhang, J., Ren, Y., Zhou, L., Fang, T., Quan, L.: Blendedmvs: A large-scale dataset for generalized multi-view stereo networks. In: IEEE/CVF Conf. Computer Vision and Pattern Recognition (CVPR). pp. 1790--1799 (2020)

\bibitem{yariv2021volume}
Yariv, L., Gu, J., Kasten, Y., Lipman, Y.: Volume rendering of neural implicit surfaces. Advances in Neural Information Processing Systems (NeurIPS)  \textbf{34},  4805--4815 (2021)

\bibitem{yariv2023bakedsdf}
Yariv, L., Hedman, P., Reiser, C., Verbin, D., Srinivasan, P.P., Szeliski, R., Barron, J.T., Mildenhall, B.: Bakedsdf: Meshing neural sdfs for real-time view synthesis. In: ACM SIGGRAPH (2023)

\bibitem{yariv2020multiview}
Yariv, L., Kasten, Y., Moran, D., Galun, M., Atzmon, M., Ronen, B., Lipman, Y.: Multiview neural surface reconstruction by disentangling geometry and appearance. Advances in Neural Information Processing Systems (NeurIPS)  \textbf{33},  2492--2502 (2020)

\bibitem{yu2022monosdf}
Yu, Z., Peng, S., Niemeyer, M., Sattler, T., Geiger, A.: Monosdf: Exploring monocular geometric cues for neural implicit surface reconstruction. Advances in Neural Information Processing Systems (NeurIPS)  \textbf{35},  25018--25032 (2022)

\bibitem{Zhang_2022_CVPR}
Zhang, J., Yao, Y., Li, S., Fang, T., McKinnon, D., Tsin, Y., Quan, L.: Critical regularizations for neural surface reconstruction in the wild. In: IEEE/CVF Conf. Computer Vision and Pattern Recognition (CVPR). pp. 6270--6279 (2022)

\bibitem{zhang2021learning}
Zhang, J., Yao, Y., Quan, L.: Learning signed distance field for multi-view surface reconstruction. In: IEEE/CVF Int'l Conf. Computer Vision (ICCV). pp. 6525--6534 (2021)

\bibitem{kaizhang2020}
Zhang, K., Riegler, G., Snavely, N., Koltun, V.: Nerf++: Analyzing and improving neural radiance fields. arXiv:2010.07492  (2020)

\bibitem{Zhang_2022_CVPR_1}
Zhang, X., Bi, S., Sunkavalli, K., Su, H., Xu, Z.: Nerfusion: Fusing radiance fields for large-scale scene reconstruction. In: IEEE/CVF Conf. Computer Vision and Pattern Recognition (CVPR). pp. 5449--5458 (2022)

\bibitem{zhang2023towards}
Zhang, Y., Hu, Z., Wu, H., Zhao, M., Li, L., Zou, Z., Fan, C.: Towards unbiased volume rendering of neural implicit surfaces with geometry priors. In: IEEE/CVF Conf. Computer Vision and Pattern Recognition (CVPR). pp. 4359--4368 (2023)

\bibitem{Zhu_2022_CVPR}
Zhu, Z., Peng, S., Larsson, V., Xu, W., Bao, H., Cui, Z., Oswald, M.R., Pollefeys, M.: Nice-slam: Neural implicit scalable encoding for slam. In: IEEE/CVF Conf. Computer Vision and Pattern Recognition (CVPR). pp. 12786--12796 (2022)

\end{thebibliography}
}%
}{}%
\end{document}